\documentclass{article}

\usepackage{arxiv}

\usepackage[utf8]{inputenc} 
\usepackage[T1]{fontenc}    
\usepackage{hyperref}       
\usepackage{url}            
\usepackage{booktabs}       
\usepackage{amsfonts}       
\usepackage{nicefrac}       
\usepackage{microtype}      
\usepackage{lipsum}		
\usepackage{graphicx}
\usepackage{natbib}
\usepackage{doi}

\usepackage[utf8]{inputenc} 
\usepackage[T1]{fontenc}    
\usepackage{hyperref}       
\usepackage{url}            
\usepackage{booktabs}       
\usepackage{amsfonts}       
\usepackage{nicefrac}       
\usepackage{microtype}      
\usepackage{xcolor}         
\usepackage{amsmath}
\usepackage{graphicx}
\usepackage[ruled, vlined]{algorithm2e}
\usepackage{algorithmic}
\usepackage{caption}
\usepackage{subcaption}
\usepackage{multirow}
\usepackage{multicol}
\usepackage{tablefootnote}
\usepackage{wrapfig}
\usepackage{textcomp}

\title{KCNet: An Insect-Inspired Single-Hidden-Layer Neural Network 
with Randomized Binary Weights for Prediction and Classification Tasks}

\date{} 					

\author{ \href{https://orcid.org/0000-0003-4429-3311}{\includegraphics[scale=0.06]{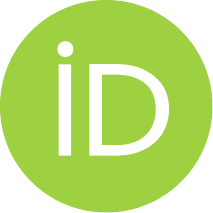}\hspace{1mm}Jinyung Hong}\\
	School of Computing and Augmented Intelligence\\
	Arizona State University\\
	Tempe, AZ 85281 \\
	\texttt{jhong53@asu.edu} \\
    \And
	\href{https://orcid.org/0000-0002-7073-6932}{\includegraphics[scale=0.06]{orcid.pdf}\hspace{1mm}Theodore P. Pavlic} \\
	School of Computing and Augmented Intelligence \\
    School of Life Sciences \\
	Arizona State University \\
    Tempe, AZ 85281 \\
	\texttt{tpavlic@asu.edu} \\
}

\renewcommand{\shorttitle}{\textit{arXiv} Template}

\hypersetup{
pdftitle={A template for the arxiv style},
pdfsubject={q-bio.NC, q-bio.QM},
pdfauthor={David S.~Hippocampus, Elias D.~Striatum},
pdfkeywords={First keyword, Second keyword, More},
}

\begin{document}
\maketitle

\begin{abstract}
Fruit flies are established model systems for studying olfactory learning as they will readily learn to associate odors with both electric shock or sugar rewards. The mechanisms of the insect brain apparently responsible for odor learning form a relatively shallow neuronal architecture. Olfactory inputs are received by the antennal lobe~(AL) of the brain, which produces an encoding of each odor mixture across $\sim$50 sub-units known as glomeruli. Each of these glomeruli then projects its component of this feature vector to several of $\sim$2000 so-called Kenyon Cells~(KCs) in a region of the brain known as the mushroom body~(MB). Fly responses to odors are generated by small downstream neutrophils that decode the higher-order representation from the MB. Research has shown that there is no recognizable pattern in the glomeruli--KC connections (and thus the particular higher-order representations); they are akin to fingerprints~-- even isogenic flies have different projections. Leveraging insights from this architecture, we propose \emph{KCNet}, a single-hidden-layer neural network that contains sparse, randomized, binary weights between the input layer and the hidden layer and analytically learned weights between the hidden layer and the output layer. Furthermore, we also propose a dynamic optimization algorithm that enables the KCNet to increase performance beyond its structural limits by searching for a more efficient set of inputs. For odorant-perception tasks that predict the perceptual properties of an odorant, we show that KCNet outperforms existing data-driven approaches, such as XGBoost. For image classification tasks, KCNet achieves reasonable performance on benchmark datasets (MNIST, Fashion-MNIST, and EMNIST) without any data-augmentation methods or convolutional layers and shows a particularly fast running time. Thus, neural networks inspired by the insect brain can be both economical and perform well.
\end{abstract}


\section{Introduction}
\label{sec:introduction}

Connectomics~\citep{seung2012connectome} aims to produce a blueprint of the connections between all neurons. Such maps promise to help illuminate how the brain learns and how memories are formed. Active connectome research focuses not only on the human brain~\citep{sporns2005human} but also on the brains of insects~\citep{eichler2017complete, takemura2017connectome}. 
The fruit fly \emph{Drosophila melanogaster} has shown excellent associative learning ability to link certain odorant stimuli~(i.e., complex mixtures of odor compounds) to rewards or penalties~\citep{caron2013brains}; the fly tends to recognize a mixture of stimuli as if they were a unitary object~\citep{deisig2001configural, devaud2015neural}. 
Studies of the fruit fly nervous system (as well as related studies in insects in general) have revealed that olfactory signals from the antenna terminate in a so-called antennal lobe~(AL) of the brain, which is a bundle of $\sim$50 discrete subunits known as glomeruli. 
These glomeruli act as feature extractors, turning each complex odor mixture into a 50-digit code. 
Projection neurons~(PNs) connect each of these glomeruli code components to multiple of the $\sim$2000 so-called Kenyon Cells~(KC), which are located in a remote area of the insect brain known as the mushroom bodies~(MB's). Thus, the projection from the AL to the MB represents a massive representational expansion. 
These KCs in the MB are then decoded by smaller neuropils that lead to the actuation of the fly behavior. From a deep neural network perspective,  this architecture has the form of a single-hidden-layer network with an input layer~(the AL), a hidden layer~(the MB), and several parallel output layers~(the downstream neuropils that control fly motion).
However, research into the topology of the AL--MB connections has
revealed that the inputs to each KC are randomized, where even isogenic flies appear to express different connections~\citep{caron2013random}. 
Furthermore, feedback inhibition within the MB regulates the excitability of the KCs to small regions, thereby enforcing a sparse representation of each odor across the MBs that, when artificially disrupted, greatly reduces the ability of flies to discriminate between odors~\citep{lin2014sparse}. 
Thus, flies achieve neuronal function through the involvement of only a small fraction of their latent-variable space in the MBs~\citep{honegger2011cellular}, where each of these latent variables is effectively a randomized combination of a few components of each feature~\citep{caron2013brains}.

Meanwhile, Artificial Neural Networks~(ANNs) and deep learning have shown excellent performance as tools applied in various fields such as computer vision, speech recognition, natural language processing, and nonlinear predictive classifiers that can extract useful features from input data~\citep{scardapane2017randomness, cao2018review}. 
However, training ANNs typically requires optimizing a non-convex 
objective function with multiple local optima and saddle 
points~\citep{glorot2010understanding}, which is difficult and costly to optimize. 
For this reason, neural networks with relatively simple structures~(i.e., single hidden layer) have gained renewed attention~\citep{cao2018review}. The architecture of these models is easy to implement and allows reasonable performance through simple optimization and fast learning processes. A promising technique for the development of such networks is \emph{Randomization}~--construction of a neural network with a subset of randomly generated and fixed weights so that the remaining weights can be learned in a straightforward linear optimization problem. By projecting inputs randomly into a high-dimensional space and then focusing training on finding weights that map this space to the output, learning can be solved as a highly tractable linear least square problem~\citep{scardapane2017randomness}. 
Furthermore, not only is the nature of the optimization problem simpler, but the number of weights that need to be trained is only a fraction of the total weights in the network. Such networks have universal approximation properties~\citep{igelnik1995stochastic, li1997comments, huang2006universal}, which has allowed them to be broadly applied to classification and regression problems.

In this paper, we propose a novel approach for generating randomized, single-hidden-layer networks based on insights from the fly brain that allows for increasing the efficiency of these simple architectures. In particular, we introduce \emph{KCNet}, a single-hidden-layer neural network with \emph{sparse} and randomized \emph{binary} weights that mimic how odor stimuli are encoded in the insect brain's high-dimensional KC representation. In addition, we also define a dynamic optimization method based on the straight-through gradient estimator~\citep{bengio2013estimating}, which allows the model to explore a more efficient set of inputs dynamically. 
We first apply our model to odorant-perception tasks, which determines how the perceptual characteristics of odorants can relate to the structural properties of the molecules. We demonstrate that our model achieves better performance compared to the existing data-driven approaches, such as XGBoost. Furthermore, to demonstrate the proposed model's versatility, we also address image classification tasks on image benchmark datasets, including MNIST, Fashion-MNIST, and EMNIST. 
We show that without the need for data augmentation or convolutional layers, KCNet can achieve reasonable performance.

\section{Related Work}
\label{sec:related_work}
Our work is strongly inspired by insect neuroscience, particularly the olfactory architecture of the fruit-fly nervous system. Olfactory sensory neurons located on the antennae and maxillary palps of fruit flies are responsible for detecting odors. The majority of these neurons only express one of the $\sim$50 odor-receptor proteins~\citep{clyne1999novel, vosshall1999spatial, couto2005molecular, fishilevich2005genetic}. The olfactory sensory neurons are projected to the AL, where they are merged so that the neurons representing the same receptor gather in one of the $\sim$50 glomeruli~\citep{couto2005molecular, fishilevich2005genetic}. Thus, this wiring diagram in which each odorant receptor is mapped to one glomerulus shows how odors are represented in the AL. When an odorant stimulus or a molecule binds to its cognate receptors, the sensory neurons expressing these receptors fire, and an odor-specific pattern of glomerular activity emerges~\citep{wang2003two}. PNs involved with each glomerulus pass olfactory information to two higher brain centers: the lateral horn and the MB. Specifically, the MB is responsible for the fly’s ability to learn to approach or avoid odors, and an input neuropil of the MB, called the \emph{calyx}, contains the axon terminals of the $\sim$50 types of PNs that connect with $\sim$2000 KCs~(i.e., a $40\times$ representational expansion). This paper mainly focuses on the transformation of odorant representation in the AL to the higher-order representation across the KCs of the MB and how to implement artificial KCs in our ANN model.

Our work is also influenced by Neural Network with Random Weights~(NNRW), a single-hidden-layer neural network with random weights from the input layer to the hidden layer. The weights between the input and the hidden layers are randomly generated in a suitable range and kept fixed during the training process. Compared with conventional learning methods with global tuning, such as deep learning with backpropagation, NNRW mitigates problems related to slow learning convergence and local optimality traps due to highly non-convex optimization.
NNRW can achieve a much faster training speed with acceptable accuracy. Furthermore, NNRW is easy to implement, and is theoretically capable of universal approximation~\citep{igelnik1995stochastic, li1997comments, huang2006universal}. Although many studies of NNRWs have been conducted under several names, such as Extreme Learning Machine~(ELM)~\citep{huang2006extreme}, Random Vector-Functional Link~(RVFL) networks~\citep{pao1992functional}, and Schmidt's method~\citep{schmidt1992feed}, the underlying ideas and training mechanism of the three methods are very similar, and so we briefly introduce the formulation of ELM model as follows.

Given a generic input $\textbf{x} \in \mathbb{R}^{d}$, the basic ELM model can be represented by linearly combining $\mathcal{B}$ random nonlinear transformations of the original input, as in
\begin{equation}
    f_{\mathcal{B}}(\textbf{x}) = \sum_{i=1}^{\mathcal{B}} \beta_{i} h_{i}(\textbf{x} ; \textbf{w}_{i}) 
    \label{eq:nnrw_all}
\end{equation}
where $i$th transformation $h_i$ is parameterized by a vector $\textbf{w}_{i}$ of randomly drawn weights from a specific probabilistic distribution or a certain rule. 
In contrast, the linear coefficients $\beta_{i}$ are trained using training data. Thus, inputs are randomly transformed into hidden, latent variables, which the output layer is trained to decode. 
In the \emph{addictive} case, a hidden node $h_{i}(\cdot)$ of the model applies a predefined nonlinearity on a random affine combination of the input vector, as in:
\begin{equation}
    h_{i}(\textbf{x}) = g(\textbf{a}_{i}^{T}\textbf{x} + b_{i})
    \label{eq:nnrw_hidden}
\end{equation}
where random vector $\textbf{w}_{i}=[\textbf{a}_{i}^{T}, b_{i}]^{T} \in \mathbb{R}^{d+1}$ and $g$ is a nonlinear function, such as a sigmoid. Given a set of $N$ data samples, where each sample $i \in \{1,\dots,N\}$ consists of an input vector $\textbf{x}_i$ and a one-hot-encoded target vector $\textbf{y}_i \in \mathbb{R}^c$, a properly trained model will best approximate the target matrix $\textbf{Y}$ with the linear system $\textbf{H} \boldsymbol \beta$, where
%
%
which can be written compactly as $\textbf{H}\boldsymbol\beta = \textbf{Y}$ where
%
%
%
\begin{gather}
    \textbf{H} = 
    \begin{pmatrix}
        h_{1}(\textbf{x}_1; \textbf{w}_{1}) & \dots & h_{\mathcal{B}}(\textbf{x}_1; \textbf{w}_{\mathcal{B}}) \\
        \vdots & \ddots & \vdots \\
        h_{1}(\textbf{x}_N; \textbf{w}_{1}) & \dots & h_{\mathcal{B}}(\textbf{x}_N; \textbf{w}_{\mathcal{B}})
    \end{pmatrix}_{N \times \mathcal{B}}
    \; \text{and} \quad
    \boldsymbol\beta = \begin{pmatrix}
        \boldsymbol\beta_{1}^{T} \\
        \vdots \\
        \boldsymbol\beta_{\mathcal{B}}^{T} \\
    \end{pmatrix}_{\mathcal{B} \times c}
    \; \text{and} \quad
    \textbf{Y} = 
    \begin{pmatrix}
        \textbf{y}_{1}^{T} \\
        \vdots \\
        \textbf{y}_{\mathcal{B}}^{T} \\
    \end{pmatrix}_{N \times c}.
    \label{eq:nnrw_matrices}
\end{gather}
That is, \textbf{H} is the hidden layer output matrix of the network; the $i$th column of \textbf{H} indicates the $i$th hidden node's output vector with respect to $\textbf{x}_{1}, \dots, \textbf{x}_{N}$, and the $j$th row of \textbf{H} represents the output vector of the hidden layer with respect to input $\textbf{x}_{j}$. Because the learning problem is equivalent to a linear regression of the output~$(\textbf{Y})$ on the transformed versions of the inputs~(embedded in $\textbf{H}$), the output weights $\boldsymbol\beta$ can be estimated as $\boldsymbol\beta^{*} = \textbf{H}^{\dagger}\textbf{Y}$
%
%
where $\textbf{H}^{\dagger}$ is the Moore--Penrose generalized inverse~\citep{banerjee1973generalized, ben2003generalized} of the hidden layer output matrix $\textbf{H}$. As discussed herein, our model is loosely connected to a specific configuration of NNRWs and exploits the learning process of training the output weights $\boldsymbol\beta$.

\section{KCNet: An Insect-Inspired Single Hidden Layer Neural Network}
\label{sec:kcnet}

\subsection{Intuition}
\label{sec:kcnet_intuition}

For bio-inspiration of our model, we focus our attention on: (i) how to form random connections between the AL glomeruli~(input layer) and the KCs~(hidden layer), and (ii) how to ensure hidden-layer sparsity to best differentiate one odor stimulus from another. To define an arbitrary set of inputs for each KC in the model, the connections between glomeruli and KCs can be represented by sparse, binary, and random matrices~\citep{caron2013random}. The firing rates from $\sim$7 randomly selected glomeruli are passed and summed to each KC~\citep{caron2013random, inada2017origins}. For example, \citet{dasgupta2017neural} proposed a computational model of the fly olfactory system, including a sparse binary random projection matrix, to generate a \emph{tag} for each odor composed of a set of neurons that are excited when that odor is presented.
We also refer to that definition and take 7 random indices of the input vector to generate the input for each KC.

The adequate method of representing the KCs' output activity is critical because the output of KCs is responsible for learning behavioral responses to different odors~\citep{owald2015olfactory}. In their fly-inspired algorithm,
\citet{dasgupta2017neural} use a Winner-Take-All~(WTA) approach that models inhibitory feedback ensuring that only the highest-firing 5\% of KCs are used on the output. Their WTA circuit selects the indices of the top $k$ values of KC activities; however, this method is not ideal for the optimization approach that we introduce later. Therefore, we base our algorithm on a computational model of the fly brain from \citet{endo2020synthesis} that generates sparsity from global inhibitory feedback between KCs~\citep{gruntman2013integration, lin2014sparse, inada2017origins}. In their model, KCs produce an intermediate output that is then subject to global inhibition from the average glomerular input to all KCs. The final KC is then generated by thresholding the inhibited output through a ramp function, which is functionally equivalent the Rectified Linear Unit (ReLU) activation function in artificial neural networks~\citep{glorot2011deep}. Thus, the output of the $j$th KC, $\mathcal{KC}_{out_{j}}$, was described as:
\begin{equation}
    \mathcal{KC}_{out_{j}} = \phi\left( \mathcal{KC}_{in_{j}} - C\frac{1}{N_{\mathcal{KC}}}\sum_{j}^{N_{\mathcal{KC}}}\mathcal{KC}_{in_{j}} \right)
\label{eq:r_out_bio}
\end{equation}
where $\mathcal{KC}_{in_{j}}$ indicates the weighted sum of input from 7 random indices of the input vector\footnote{See precise weighting details from \citet{endo2020synthesis}.}, $\phi$ is the ReLU, $C$ is the strength of global inhibition, and $N_{\mathcal{KC}}$ is the total number of KCs. Herein, we use $C=1.0$ and $N_{\mathcal{KC}}=2000$ matching the values best calibrated to real KC responses~\citep{endo2020synthesis, aso2009mushroom}.

\subsection{Model Architecture}
\label{sec:kcnet_architecture}

\begin{figure}\centering
    \begin{subfigure}[b]{0.4\textwidth}\centering
        \includegraphics[width=\textwidth]{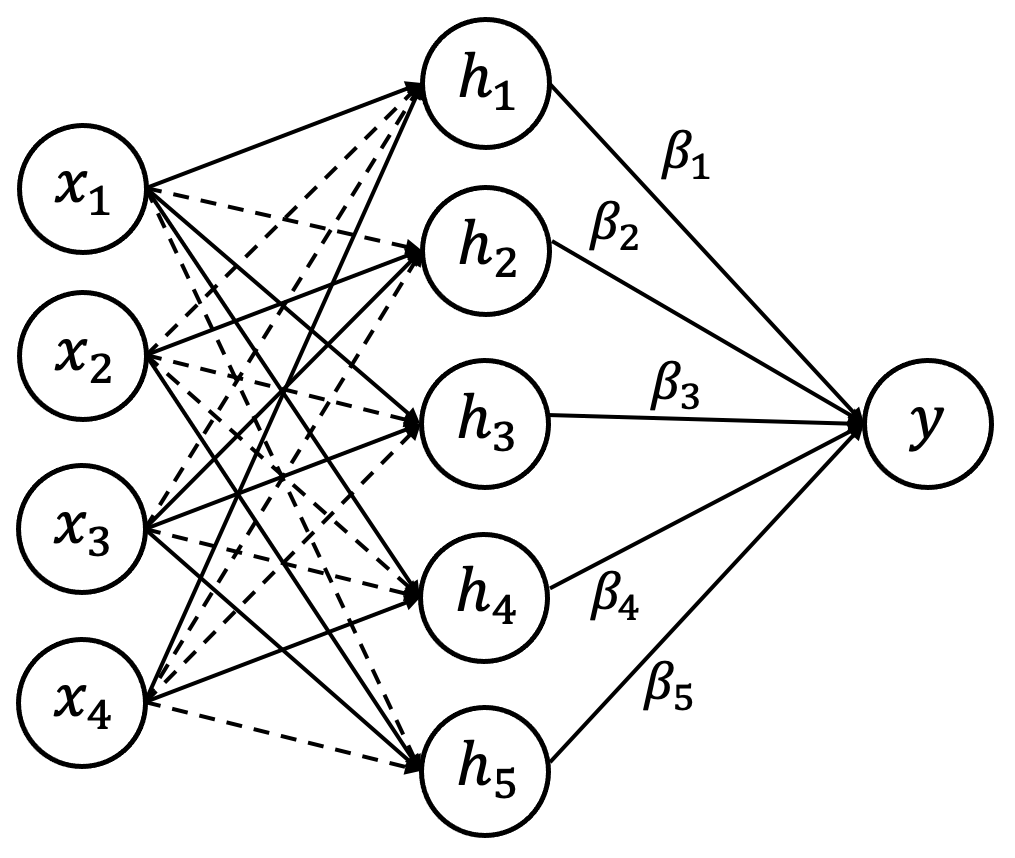}
        \caption{Overall architecture of the KCNet. Between the input layer and the hidden layer, solid~(dashed) lines indicate weight of~1~(0).}
        \label{fig:kcnet_overall}
    \end{subfigure}
    \hfill
    \begin{subfigure}[b]{0.52\textwidth}\centering
        \includegraphics[width=\textwidth]{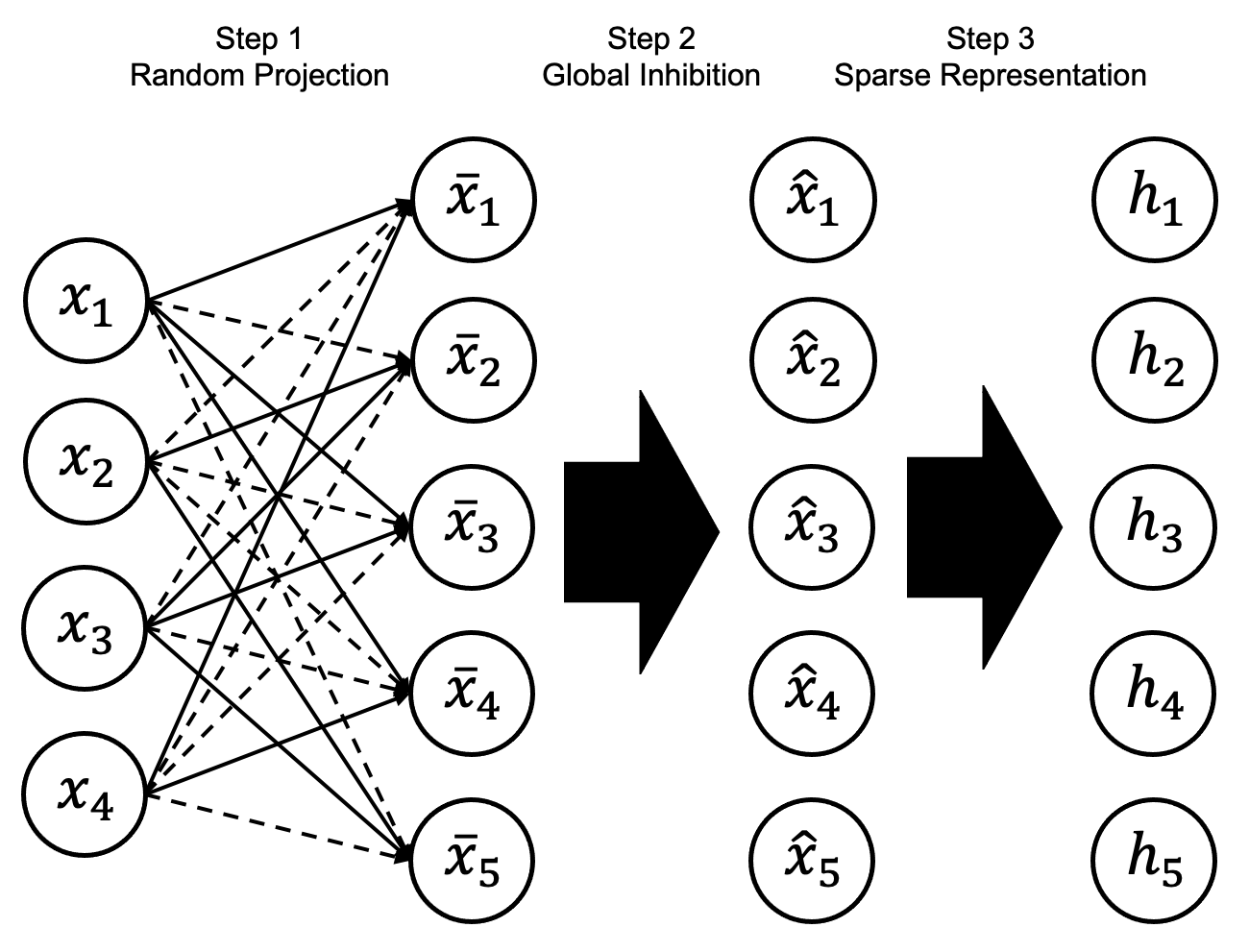}
        \caption{Process of producing hidden layer output. Step 1 is random projection, Step 2 in Eq.~(\ref{eq:kcnet_intermediate}), and Step 3 in Eq.~(\ref{eq:kcnet_output}).} 
        \label{fig:kcnet_process}
    \end{subfigure}
\caption{Visualization of model architecture of the KCNet}
\label{fig:kcnet_architecture}
\end{figure}

We take input from the feature space $\mathcal{X} \subseteq \mathbb{R}^d$ and derive our model for a single output for simplicity. Our proposed model with $\mathcal{B}$ ($\sim N_{\mathcal{KC}}$) hidden nodes is a parametric function $f_{\mathcal{B}}: \mathbb{R}^{d} \to \mathbb{R}$ taking an input $\textbf{x} = [x_1, x_2, \ldots, x_d]^{T}$. To define our randomized encoder in the hidden nodes, we select $N_{in} \in [1,d)$ indices of the input at random without replacement for each hidden node\footnote{$N_{in} < d$ to prevent hidden-unit outputs becoming trivially 0 by Eq.~(\ref{eq:kcnet_intermediate}). In our setting, $N_{in}=7$.}. In other words, the output $\Bar{x}_j$ of each hidden node $j \in \{1,...,\mathcal{B}\}$ is a weighted combination of all $d$ inputs where only $N_{in} < d$ inputs have $w_{j,i}=1$ and all other inputs have $w_{j,i}=0$. So the inputs are effectively gated by the weights on each hidden node.
%
%
Once the $j$th hidden node has produced weighted sum $\Bar{x}_{j}$, the post-processing step to produce intermediate output is performed in the hidden layer. Mimicking Eq.~(\ref{eq:r_out_bio}) with $C=1$, the $j$th intermediate output $\hat{x}_{j}$ is:
\begin{equation}
    \hat{x}_{j} = \Bar{x}_{j} - \mu, \;\; \mu = \frac{1}{\mathcal{B}}\sum_{i=1}^{\mathcal{B}} \Bar{x}_{i}
    \label{eq:kcnet_intermediate}
\end{equation}
where $\mathcal{B}$ is the number of hidden units. In the model's decoder definition, the proposed network is defined as follows in the same way that a linear model is applied as a decoder in NNRW:
\begin{equation}
    f_{\mathcal{B}}(\textbf{x)} = \sum_{i=1}^{\mathcal{B}} \beta_{i}h_{i} = \sum_{i=1}^{\mathcal{B}} \beta_{i}g(\hat{x}_{i})
\label{eq:kcnet_output}
\end{equation}
where $g$ is the ReLU function~\citep{glorot2011deep} that allows the model to produce sparse hidden output, which is more biologically plausible. Therefore, our model can be formulated in the same form as in NNRW with the hidden layer output matrix \textbf{H} and target matrix \textbf{Y} from Eq.~(\ref{eq:nnrw_matrices}). However, in contrast with NNRW, we estimate $\boldsymbol\beta$ with a regularized Moore--Penrose inverse:
\begin{equation}
    \boldsymbol\beta^{*} = (\textbf{H}^{T}\textbf{H} + \lambda \textbf{I})^{-1}\textbf{H}^{T}\textbf{Y}
    \label{eq:kcnet_beta_estimation}
\end{equation}
where $\lambda$ is a specific regularization term; using a ridge regularization helps to improve generalizability and performance of the model. The overall architecture of the KCNet is depicted in Fig.~\ref{fig:kcnet_architecture}.
Our model has several noticeable characteristics that can be translated in terms of computer science and machine learning, and the details can be seen in Appendix~\ref{ap:prop_kcnet}.

\subsection{Dynamic Optimization Algorithm~(DOA) For Feature Selection} 
\label{sec:kcnet_doa}

Although KCNet can perform well by increasing the number of the nodes in the hidden layer (i.e., the hidden layer's width), we define the Dynamic Optimization Algorithm~(DOA), an iterative method that allows the KCNet to explore a more efficient set of random inputs to obtain better performance than conventional ones. For a fully trained single-hidden-layer neural network, all the hidden weights and the output weights are optimized with respect to a dataset and a loss function using gradient descent, whereas the hidden weights of the KCNet are initially fixed to either 0 or 1, and training adapts only the output weights during its learning process. What if we can find efficient combinations of inputs that allow the model to achieve better performance? How can we find which connection is effective while turning on or off the connection between the input and hidden layers? Our algorithm never updates the weights directly to address the issue. Since it associates two options, 0 or 1, for each connection, we instead optimize the \emph{preference} to select the weight between 0 and 1 to use.

\paragraph{Initialization.}
Let $w_{ji}$ be the connection between the $i$th input node and the $j$th hidden node, and let $s_{ji} \in [-1, 1]$ be the \emph{preference score} for $w_{ji}$, indicating the degree to which it prefers to choose between 0 and 1. We initially draw $s_{ji} \sim \mathbb{U}(lb, ub)$ randomly from a uniform distribution between $lb=-1$ and $ub=1$ after generating the randomized binary weight $w_{ji} \in \{0,1\}$ on each input. Then, we can readily convert the weight $w_{ji}$ into the preference score $s_{ji}$, and vice versa, by defining selection functions $\rho_{w \rightarrow s}$ and $\rho_{s \rightarrow w}$ as follows:
\begin{gather}
    s_{ji} = \rho_{w \rightarrow s}(w_{ji}) \sim
    \begin{cases}
        \mathbb{U}(0, ub) &w_{ji} = 1, \\
        \mathbb{U}(lb, 0) &\text{otherwise}
    \end{cases}
    \quad\; \text{and} \;\quad
    w_{ji} = \rho_{s \rightarrow w}(s_{ji}) =
    \begin{cases}
        1 &s_{ji} > 0, \\
        0 &\text{otherwise}
    \end{cases}
    \label{eq:selection_func_s_w}
\end{gather}

With the above setting, we can start by iterating the following forward pass and the backward pass to update the preference score $s_{ij}$ instead of $w_{ij}$ using the gradient descent method. However, because neither selection functions $\rho$ are differentiable, we cannot directly compute the gradients for $s_{ji}$. We explain how to solve this in the part of the backward pass.

\paragraph{Forward Pass: Training Output Weight.}
We split the dataset into two sets randomly; the training set for learning the output weight $\beta$ and the validation set for searching a random collection of input. Then, we build a model using the weight $w_{ji}$ and train $\beta$ using a least-squares method.


\paragraph{Backward Pass: Updating Preference Score.}
In the backward pass, all the preference scores are updated with gradient descent method based on the straight-through gradient estimation~\citep{bengio2013estimating} using the validation set by propagating to $s_{ji}$ the gradients that $w_{ji}$ receives. Specifically, the selection function $\rho_{s \rightarrow w}$ is a kind of threshold function in Eq.~(\ref{eq:selection_func_s_w}), and we ignore the derivative of the function and pass on the gradients that $w_{ji}$ receives as if the function was an identity function. Therefore, we can bypass the selection function in the backward pass and compute the loss gradient with respect to $s_{ji}$:
\begin{equation}
    \nabla_{s_{ji}}
    = \partial \mathcal{L}/\partial s_{ji}
    \approx \partial \mathcal{L} / \partial w_{ji}
    \label{eq:kcnet_backprop}
\end{equation}
where $\mathcal{L}$ is the objective function. Therefore, it is then back-propagated to obtain gradients on the parameters that influence $s_{ji}$ (see the detail of the full derivative in Appendix~\ref{ap:derivative}).
Similar to usual back-propagation method, the preference score $s_{ji}$ is updated by gradient descent with learning rate $\alpha > 0$:
\begin{equation}
    s_{ji} \leftarrow s_{ji} - \alpha\nabla_{s_{ji}}
\end{equation}
and then the updated preference score is saturated to be between $-1$ and $1$. Based on the new score, the weight $w_{ji}$ is also updated by applying the selection function $\rho_{s \rightarrow w}$ to it, and the forward pass to train $\beta$ of the model performs again. We call the pair of forward and backward passes an \emph{epoch}, and by continually iterating epochs, the KCNet can automatically find a more efficient combination of inputs that distinguish the representation belonging to one category from others. To terminate the iterative process, we established two stopping criteria:
i) if a process has run a certain number of times, and
ii) if the performance of the current model exceeds the standard we set.
Our experiments demonstrate that this simple algorithm allows the model to find convincing configurations of random inputs resulting in good results across different datasets (see details in Algorithm~\ref{alg:doa} in Appendix).

\paragraph{Ensemble DOA to Assemble a Weight Matrix.}
Although the above DOA can enhance the conventional KCNet, the computational burden drastically increases if the size of the dataset or the number of hidden nodes in the model is enormous due to, for example, the computation of an inverse matrix or an inner product between input matrix and the derivative of a loss matrix with respect to the input in the backward pass.
Furthermore, we noticed that the larger the size of the weight matrix, the more perturbed the convergence of the performance of the DOA-applied KCNet and the more sensitive the learning rate (see Fig.~\ref{fig:relation_hsize_lr} in Appendix). Thus, instead of directly applying the DOA to the KCNet with a large weight matrix, we introduce another use case of DOA, \emph{Ensemble DOA}, to assemble a large weight matrix by concatenating several small weight matrices after each small weight matrix was adapted individually by the DOA. In our experiments, we empirically determined that it was most effective to divide a large weight matrix into more than 10 small matrices. In this condition, KCNet with the Ensemble DOA shows a much faster training speed than the KCNet with the DOA and outperforms the original KCNet and the KCNet directly applied by the DOA.

\section{Experiments}
\label{sec:exp}

\subsection{Odor Perception}
The odorant perception task is the prediction of the identity of an odor from the physical, chemical, and structural properties of its odor molecules. Understanding odorant perception continues to be an active area of research due to the complexity of the task. First of all, the dimensionality of the feature space for odorant perception and the relationship between the structure and odor are still unknown~\citep{bushdid2014humans, mamlouk2004dimensions, sell2006unpredictability}. Furthermore, odor molecules cannot be predicted to have the same odor properties by looking at similarities in structural properties~\citep{bentley2006nose,egawa2006structural}. Rather, odor molecules with completely different structural features may have similar scents~\citep{amoore1970molecular}. And depending on how concentrated the odor stimulus is in the environment, it can be perceived in many different smells~\citep{amoore1970molecular}.

\Citet{chacko2020data} focused on a specialized form of the odorant perception task where odors could be classified by the presence or absence of two particular subjective properties referred to as \emph{Sweet} and \emph{Musky}. They used a psychophysical dataset including odor perception ratings of 55 human subjects. Based on this dataset, \citeauthor{chacko2020data} used machine learning to predict these perceptual characteristics from an odor molecule based on its structural properties. We applied our models to the same dataset, and we found that our KCNet models outperform the other data-driven approaches, such as XGBoost, that were used by \citet{chacko2020data}.

\paragraph{Data Pre-processing.}
Here, we briefly summarize the data pre-processing step used by \citet{chacko2020data} to define the odor-perception classification problem. For input labels, \citet{keller2016olfactory} proposed a dataset that consists of odorant attribute data for 480 structurally distinct compounds at two concentrations. The odor properties included were 20 semantic descriptors, such as \emph{Sweet, Musky, Flower}, and so on, associating with odor quality.
55 subjects evaluated each of these perceptual attributes in the range of 0 to 100. The degree of the evaluation indicates how well each semantic descriptor describes the perceptual characteristics of the odor.  We assigned ground-truth labels for their odor quality to each odorant compound based on the frequency of using semantic descriptors assessed by the 55 subjects. To consider the complexity of odor perception and the possibility that an odor stimulus can be recognized to have multiple odor characteristics, we selected the top three semantic descriptors for each odorant compound. Thus, we defined each odor stimulus to be represented by three Odor Characters~(OCs).

For input features, we utilized 196 two-dimensional RDKit molecular data as input features~\citep{hall1991molecular}. It consists of topological descriptors, connectivity descriptors, constitutional descriptors, molecular property, and MOE-type descriptors and provides quantitative information about the physical, chemical, and topological features measured in the 2D graph representations of the structures. After some zero values of the descriptors were removed and the dilution was converted into the one-hot encoding as a categorical feature, we have 960 data samples (480 compounds $\times$ 2 dilutions) with 159 input features.
After data pre-processing, we noticed that other OCs showed a high imbalance in their distribution, whereas \emph{Sweet} and \emph{Musky} OCs had enough samples for the positive class (58.13\%, 41.9\%, respectively). Therefore, we formulated it as a binary classification task by creating two datasets: one for the \emph{Sweet} OC and another for the \emph{Musky} OC.

\paragraph{Experimental Detail.}

We randomly split the processed dataset into a training and testing set with a 9:1 ratio after applying a standard normalization. Only the training set was utilized for model parameter tuning using a 5-fold cross-validation.
During cross-validation, the training set was split into five folds at every iteration, where one fold was kept as a validation set to test the performance with the trained parameters. 
For the classification of \emph{Musky} OC, we applied an oversampling method to the training data, excluding the validation set, to reduce the imbalance in the data distribution. The weighted F1 score was used as an evaluation metric because of the imbalanced characteristic of the datasets. 
We configured the hyperparameter search space consistent with the models of \citet{chacko2020data} and a specific range for the regularization term $\lambda$ for our model. We utilized the Optuna framework~\citep{akiba2019optuna} by repeating the cross-validation 100 times.
After finding the best-performing number of hidden nodes of the KCNet, we set it as the number of hidden nodes for both the DOA-applied model and the Ensemble-DOA-applied model, and we used the same value of the regularization term $\lambda$ as was used for KCNet. Details of the methods, hyperparameter search spaces, and implementations are given in Appendix~\ref{ap:exp}.
\begin{wrapfigure}{r}{0.425\textwidth}\centering%
    \includegraphics[width=0.425\textwidth]{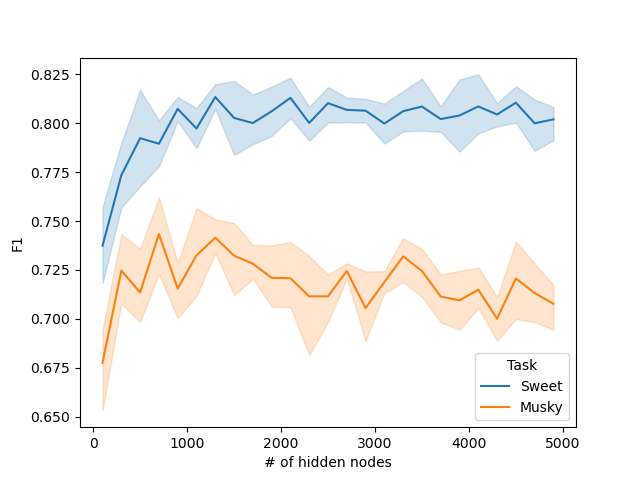}%
    \caption{Test F1 score with 95\% confidence interval of the KCNet for odor perception depending on its hidden size.}%
    \label{fig:results_odor}
\end{wrapfigure}

\paragraph{Results.}
Table~\ref{tab:result_odor} shows the performance comparison of test-weighted F1 score with 95\% confidence interval averaged by five experiments among models. The performance variances of the KCNet and its variants are natural because their weight matrices are stochastically determined, while the performances of other models are stable after their hyperparameters are fixed. Our model and its variants usually outperforms other models, and the best performing algorithm was the KCNet with Ensemble DOA for \emph{Sweet} OC and the original KCNet for \emph{Musky} OC, with test-weighted F1-scores of 0.8193 and 0.7434, respectively.
Applying the DOA to the KCNet deteriorated the performance of the KCNet for \emph{Musky} OC, which may indicate that the sample size for this class is not large enough, and the oversampling techniques might disturb the performance of the DOA. 
As shown in Fig.~\ref{fig:results_odor}, after sufficiently stretching the hidden nodes, the performance of KCNet for \emph{Sweet} OC seems stable, but the performance of KCNet for \emph{Musky} OC is more variable. However, the KCNet with the Ensemble DOA alleviated the situation and achieved better performance than the KCNet with the DOA. See the details of other plots and results in Appendix~\ref{ap:exp}.
\begin{table}
  \caption{\emph{Test-weighted F1 score of all models for odor
  perception.} ``w/ DOA'' is the DOA-applied KCNet and ``w/ Ensemble DOA''
  is the Ensemble-DOA-applied KCNet. $\text{Mean}_{\pm \text{SD}}
  (\text{\# of hidden units})$ for ``KCNet'' and ``w/ DOA'' and
  $\text{Mean}_{\pm \text{SD}} (\text{\# of submodels} \times \text{\#
  of hidden units for each submodel})$ for ``w/ Ensemble DOA''. The best
  performance is displayed in bold.}
  \centering
  \begin{tabular}{lll}
    \toprule
    & \multicolumn{2}{c}{Classification Tasks} \\
    \cmidrule(l){2-3}
    Model     & Sweet/Non-sweet     & Musky/Non-musky \\
    \midrule
    Adaboost & 0.7444  & 0.6436 \\
    Gradient Boosting Machine & 0.7538  & 0.7186 \\
    XGBoost & 0.7316  & 0.6627 \\
    Random Forest & 0.7427  & 0.6826 \\
    K-Nearest Neighbors & 0.7839  & 0.6538 \\
    Support Vector Machine & 0.7980 & 0.6629 \\
    \midrule
    KCNet &  $0.8134_{\pm 0.017} \; (\text{1,300})$  & $\textbf{0.7434}_{\pm \textbf{0.052}} \; (700) $ \\
    w/ DOA &  $0.8156_{\pm 0.067} \; (\text{1,300})$   & $0.7004_{\pm 0.027} \; (700)$ \\
    w/ Ensemble DOA &  $\textbf{0.8193}_{\pm \textbf{0.022}} \; (13 \times 100)$   & $0.7247_{\pm 0.055} \; (10 \times 70)$ \\
    \bottomrule
  \end{tabular}
  \label{tab:result_odor}
\end{table}

\subsection{Image Classification}
To show the general applicability of our model, we applied it to the image-classification task using well-known image benchmark datasets, MNIST~\citep{lecun1998gradient}, Fashion-MNIST~\citep{xiao2017fashion}, and EMNIST~\citep{cohen2017emnist}. 

\paragraph{Datasets.}
MNIST is comprised of 10-class $28 \times 28$ pixel grayscale images of handwritten digits. It contains 60,000 images for training and 10,000 images for testing. Fashion-MNIST is a dataset including $28 \times 28$ grayscale images of 70,000 fashion products from 10 categories, with 7,000 images per category. The training~(test) set has 60,000~(10,000) images. The EMNIST dataset is a set of handwritten characters and digits derived from the NIST Special Database 19 and converted to a $28 \times 28$ image format. Six datasets that comprise the EMNIST dataset but we focus one subset, EMNIST-Balanced, which has 47 classes and 131,600 images so that 112,800 images for training and 18,800 images for testing can be used. 

\paragraph{Experimental Detail.}
We implemented the KCNet with PyTorch~\citep{paszke2017automatic} to boost the training speed supported by GPUs; PyTorch supports the above three datasets so that they can be easily used. Therefore, we conducted our experiments after only applying a standard normalization to them without the use of any data-augmentation method. We also implement the ELM model and a Fully-trained Single-Hidden-layer Neural network~(FSHN) and test them to compare the performance with our model. The ELM was introduced as a baseline classifier by \citet{cohen2017emnist}, and its characteristics are very similar to our model. We formulated the ELM model in a method that is often used in practice~\citep{scardapane2017randomness, huang2006extreme, cao2018review} as follows: i) its hidden weights $\textbf{w}_{i} = [\textbf{a}_{i}^{T}, b_{i}]^{T}$ were defined by drawing uniform random variables $\textbf{a}_{i} \sim \mathbb{U}(-1,1)$ and $b_{i} \sim \mathbb{U}(0,1)$; ii) the sigmoid function is used as a nonlinearity function $g$; and iii) the output weight $\beta$ was estimated using the Moore--Penrose inverse. FSHN was implemented using the linear models followed by the ReLU function and optimized using the stochastic gradient descent, supported by PyTorch. The number of hidden nodes in the FSHN was set to have the number of adaptable parameters similar to KCNet. For hyperparameter setting in the DOA-applied KCNet and the Ensemble-DOA-applied KCNet, the original training set was randomly split into a training set and a validation set with a 5:1 ratio every epoch. See Appendix~\ref{ap:exp} for details of methods, hyperparameter search spaces, and implementations.

\paragraph{Results.} Table~\ref{tab:result_image} shows the results of test accuracy with 95\% confidence interval on the image benchmark datasets over five experiments.
\begin{table}
  \caption{\emph{Test accuracy of all models for image classification.}
  $\text{Mean}_{\pm \text{SD}} (\text{\# of hidden units})$ for ``ELM'',
  ``KCNet'' and ``w/ DOA'' and $\text{Mean}_{\pm \text{SD}} (\text{\# of
  learnable parameters})$ for ``FSHN''. $\text{Mean}_{\pm \text{SD}}
  (\text{\# of submodels} \times \text{\# of hidden units for each
  submodel})$ for ``w/ Ensemble DOA''. The best performance is displayed in bold.}
  \label{tab:result_image}
  \centering
  \resizebox{\columnwidth}{!}{\begin{tabular}{llll}
    \toprule
    & \multicolumn{3}{c}{Image Dataset} \\
    \cmidrule(l){2-4}
    Model & MNIST & Fashion-MNIST & EMNIST-Balanced \\
    \midrule
    ELM~\citep{huang2006extreme} &  $0.9658_{\pm 0.002} \; (\text{6,500})$  & $0.8813_{\pm 0.006} \; (\text{7,000})$ & $0.7747_{\pm 0.002} \; (\text{4,500})$ \\
    FSHN &  $0.975_{\pm 0.002} \; (\text{65,200})$ & $0.8671_{\pm 0.002} \; (\text{70,765})$ & $\textbf{0.8179}_{\pm \textbf{0.002}} \; (\text{211,375})$ \\
    \midrule
    KCNet &  $0.9735_{\pm 0.003} \; (\text{6,500})$  & $0.8849_{\pm 0.002} \; (\text{7,000})$ & $0.7762_{\pm 0.002} \; (\text{4,500})$ \\
    w/ DOA &  $0.9731_{\pm 0.002} \; (\text{6,500})$  & $0.8845_{\pm 0.002} \; (\text{7,000})$ & $0.7761_{\pm 0.002} \; (\text{4,500})$ \\
    w/ Ensemble DOA &  $\textbf{0.9776}_{\pm \textbf{0.001}} \; (10\times650)$  & $\textbf{0.886}_{\pm \textbf{0.001}} \; (10\times700)$ & $0.8058_{\pm 0.002} \; (10\times450)$ \\
    \bottomrule
  \end{tabular}}
\end{table}
Our model outperforms ELM, and Fig.~\ref{fig:mnist_kcnet_elm_hsize} (and Fig.~\ref{ap:fashion_mnist_emnist_kcnet_elm_hsize} in Appendix) confirms that the KCNet performs better than the ELM model across all the image benchmark datasets. 
\begin{figure*}\centering
    \begin{subfigure}[b]{0.45\textwidth}\centering
        \includegraphics[width=\textwidth]{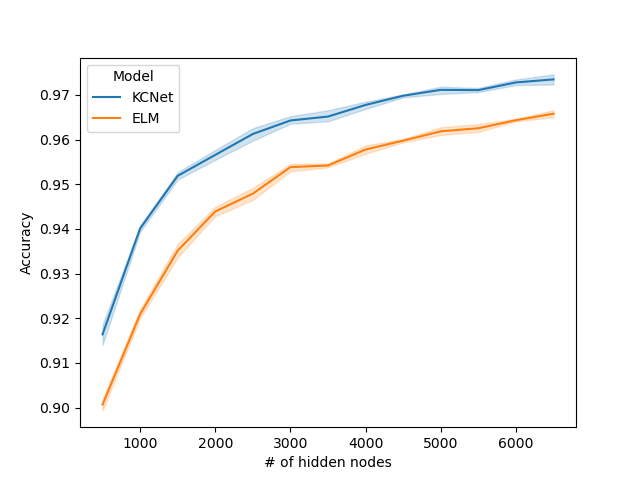}
        \caption{Test accuracy of KCNet and ELM for MNIST depending on their hidden sizes\\~}
        \label{fig:mnist_kcnet_elm_hsize}
    \end{subfigure}
    \hfill
    \begin{subfigure}[b]{0.45\textwidth}\centering
        \includegraphics[width=\textwidth]{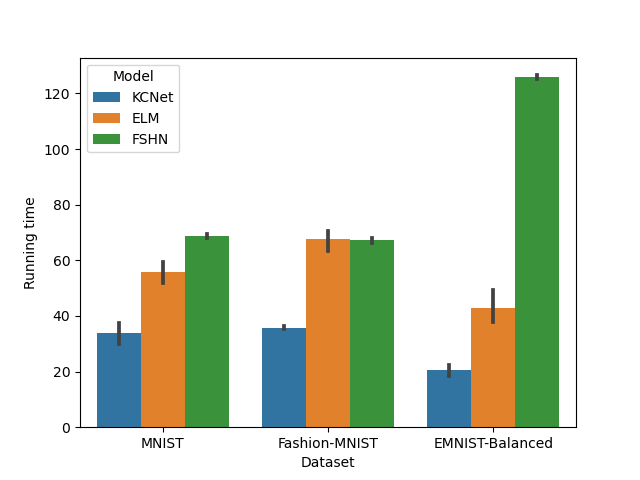}
        \caption{The comparison of running time, including model configuration time, training time, and evaluation time (sec) for KCNet, ELM, and FSHN}
        \label{fig:running_time_kcnet_elm_fshn}
    \end{subfigure}
    \hfill
    \caption{For image classification task using MNIST, the performance comparison between KCNet and ELM and the running time comparison between KCNet and FSHN}
    \label{fig:change_weight}
\end{figure*}
The best-performing model is the FSHN for EMNIST-Balanced, and the Ensemble-DOA-applied KCNet for MNIST and Fashion-MNIST. The direct application of DOA to KCNet seems rather performance-deprecating, which shows that the larger the weight matrix, the less fully exploitable the DOA. However, the Ensemble DOA is very effective at enhancing the KCNet because the DOA can be fully utilized to explore an effective set of inputs for relatively small weight matrices. Even though the performance of KCNet was worse than the FSHN, the running time of our model was much faster than FSHN (and ELM) across all the benchmark datasets as shown in Fig.~\ref{fig:running_time_kcnet_elm_fshn}. Therefore, the KCNet allows for compromising performance slightly in order to increase speed. For more details, see Appendix~\ref{ap:exp}.

\section{Conclusions and Broader Impact}
We proposed a fast, insect-brain-inspired single-hidden-layer neural network and a dynamic optimization algorithm that allows the model to explore meaningful connections in hidden weights and to utilize an ensemble method of assembling weight matrices. Although we showed the potential of our model through the above experimental results, there are several limitations yet to be addressed, as detailed in Appendix~\ref{ap:limit}. For future work, by using the discriminating power of our model, we will extend the application to other tasks, such as one-shot learning or contrastive learning. Furthermore, we will construct a more brain-like model by combining our model with biologically plausible learning algorithms, such as target propagation~\citep{meulemans2020theoretical}. 

\paragraph{Broader Impact.}
This study integrates ideas from computational insect neuroscience with machine learning to show how neural networks can be made smaller and simpler to train with little compromise in performance. The net societal impacts of this research are primarily positive. By helping to reduce the resource footprint and increase the speed of neural networks used for common tasks, this research increases the sustainability of artificial neural networks. That said, reducing the barriers to the application of AI can lead to faster growth in the use of AI, which can be positive or negative. Beyond the technological impacts, this research has positive scientific value as it further illuminates the adaptive function underlying architectures in the insect brain, potentially aiding in understanding the evolution of these architectures. Despite the many qualitative large-scale differences between arthropod and vertebrate brains, a number of similar circuit motifs and smaller-scale brain structures have been found. Thus, better understanding how insects learn to discriminate among complex signals can eventually also aid in understanding the mechanisms and evolution of similar functions in the human brain.

\section*{Acknowledgment}
This work was supported in part by NSF grant SES-1735579.

\bibliographystyle{apalike}
\bibliography{preprint_arXiv/preprint_arXiv}

\begin{thebibliography}{}

\bibitem[Akiba et~al., 2019]{akiba2019optuna}
Akiba, T., Sano, S., Yanase, T., Ohta, T., and Koyama, M. (2019).
\newblock Optuna: A next-generation hyperparameter optimization framework.
\newblock In {\em Proceedings of the 25th ACM SIGKDD international conference
  on knowledge discovery \& data mining}, pages 2623--2631.

\bibitem[Amoore, 1970]{amoore1970molecular}
Amoore, J.~E. (1970).
\newblock {\em Molecular Basis of Odor}.
\newblock Charles C.~Thomas, Springfield, Illinois.

\bibitem[Andoni and Indyk, 2006]{andoni2006near}
Andoni, A. and Indyk, P. (2006).
\newblock Near-optimal hashing algorithms for approximate nearest neighbor in
  high dimensions.
\newblock In {\em 2006 47th annual IEEE symposium on foundations of computer
  science (FOCS'06)}, pages 459--468. IEEE.

\bibitem[Aso et~al., 2009]{aso2009mushroom}
Aso, Y., Gr{\"u}bel, K., Busch, S., Friedrich, A.~B., Siwanowicz, I., and
  Tanimoto, H. (2009).
\newblock The mushroom body of adult drosophila characterized by gal4 drivers.
\newblock {\em Journal of neurogenetics}, 23(1-2):156--172.

\bibitem[Ba et~al., 2016]{ba2016layer}
Ba, J.~L., Kiros, J.~R., and Hinton, G.~E. (2016).
\newblock Layer normalization.
\newblock {\em arXiv preprint arXiv:1607.06450}.

\bibitem[Banerjee, 1973]{banerjee1973generalized}
Banerjee, K. (1973).
\newblock Generalized inverse of matrices and its applications.

\bibitem[Ben-Israel and Greville, 2003]{ben2003generalized}
Ben-Israel, A. and Greville, T.~N. (2003).
\newblock {\em Generalized inverses: theory and applications}, volume~15.
\newblock Springer Science \& Business Media.

\bibitem[Bengio et~al., 2013]{bengio2013estimating}
Bengio, Y., L{\'e}onard, N., and Courville, A. (2013).
\newblock Estimating or propagating gradients through stochastic neurons for
  conditional computation.
\newblock {\em arXiv preprint arXiv:1308.3432}.

\bibitem[Bentley, 2006]{bentley2006nose}
Bentley, R. (2006).
\newblock The nose as a stereochemist. enantiomers and odor.
\newblock {\em Chemical reviews}, 106(9):4099--4112.

\bibitem[Bushdid et~al., 2014]{bushdid2014humans}
Bushdid, C., Magnasco, M.~O., Vosshall, L.~B., and Keller, A. (2014).
\newblock Humans can discriminate more than 1 trillion olfactory stimuli.
\newblock {\em Science}, 343(6177):1370--1372.

\bibitem[Cao et~al., 2018]{cao2018review}
Cao, W., Wang, X., Ming, Z., and Gao, J. (2018).
\newblock A review on neural networks with random weights.
\newblock {\em Neurocomputing}, 275:278--287.

\bibitem[Caron, 2013]{caron2013brains}
Caron, S.~J. (2013).
\newblock Brains don't play dice—or do they?
\newblock {\em Science}, 342(6158):574--574.

\bibitem[Caron et~al., 2013]{caron2013random}
Caron, S.~J., Ruta, V., Abbott, L., and Axel, R. (2013).
\newblock Random convergence of olfactory inputs in the drosophila mushroom
  body.
\newblock {\em Nature}, 497(7447):113--117.

\bibitem[Chacko et~al., 2020]{chacko2020data}
Chacko, R., Jain, D., Patwardhan, M., Puri, A., Karande, S., and Rai, B.
  (2020).
\newblock Data based predictive models for odor perception.
\newblock {\em Scientific reports}, 10(1):1--13.

\bibitem[Clyne et~al., 1999]{clyne1999novel}
Clyne, P.~J., Warr, C.~G., Freeman, M.~R., Lessing, D., Kim, J., and Carlson,
  J.~R. (1999).
\newblock A novel family of divergent seven-transmembrane proteins: candidate
  odorant receptors in drosophila.
\newblock {\em Neuron}, 22(2):327--338.

\bibitem[Cohen et~al., 2017]{cohen2017emnist}
Cohen, G., Afshar, S., Tapson, J., and Van~Schaik, A. (2017).
\newblock Emnist: Extending mnist to handwritten letters.
\newblock In {\em 2017 International Joint Conference on Neural Networks
  (IJCNN)}, pages 2921--2926. IEEE.

\bibitem[Couto et~al., 2005]{couto2005molecular}
Couto, A., Alenius, M., and Dickson, B.~J. (2005).
\newblock Molecular, anatomical, and functional organization of the drosophila
  olfactory system.
\newblock {\em Current Biology}, 15(17):1535--1547.

\bibitem[Dasgupta et~al., 2017]{dasgupta2017neural}
Dasgupta, S., Stevens, C.~F., and Navlakha, S. (2017).
\newblock A neural algorithm for a fundamental computing problem.
\newblock {\em Science}, 358(6364):793--796.

\bibitem[Deisig et~al., 2001]{deisig2001configural}
Deisig, N., Lachnit, H., Giurfa, M., and Hellstern, F. (2001).
\newblock Configural olfactory learning in honeybees: negative and positive
  patterning discrimination.
\newblock {\em Learning \& Memory}, 8(2):70--78.

\bibitem[Devaud et~al., 2015]{devaud2015neural}
Devaud, J.-M., Papouin, T., Carcaud, J., Sandoz, J.-C., Gr{\"u}newald, B., and
  Giurfa, M. (2015).
\newblock Neural substrate for higher-order learning in an insect: mushroom
  bodies are necessary for configural discriminations.
\newblock {\em Proceedings of the National Academy of Sciences},
  112(43):E5854--E5862.

\bibitem[Egawa et~al., 2006]{egawa2006structural}
Egawa, T., Kameyama, A., and Takeuchi, H. (2006).
\newblock Structural determination of vanillin, isovanillin and ethylvanillin
  by means of gas electron diffraction and theoretical calculations.
\newblock {\em Journal of Molecular Structure}, 794(1-3):92--102.

\bibitem[Eichler et~al., 2017]{eichler2017complete}
Eichler, K., Li, F., Litwin-Kumar, A., Park, Y., Andrade, I., Schneider-Mizell,
  C.~M., Saumweber, T., Huser, A., Eschbach, C., Gerber, B., et~al. (2017).
\newblock The complete connectome of a learning and memory centre in an insect
  brain.
\newblock {\em Nature}, 548(7666):175--182.

\bibitem[Endo et~al., 2020]{endo2020synthesis}
Endo, K., Tsuchimoto, Y., and Kazama, H. (2020).
\newblock Synthesis of conserved odor object representations in a random,
  divergent-convergent network.
\newblock {\em Neuron}, 108(2):367--381.

\bibitem[Fishilevich and Vosshall, 2005]{fishilevich2005genetic}
Fishilevich, E. and Vosshall, L.~B. (2005).
\newblock Genetic and functional subdivision of the drosophila antennal lobe.
\newblock {\em Current Biology}, 15(17):1548--1553.

\bibitem[Glorot and Bengio, 2010]{glorot2010understanding}
Glorot, X. and Bengio, Y. (2010).
\newblock Understanding the difficulty of training deep feedforward neural
  networks.
\newblock In {\em Proceedings of the thirteenth international conference on
  artificial intelligence and statistics}, pages 249--256. JMLR Workshop and
  Conference Proceedings.

\bibitem[Glorot et~al., 2011]{glorot2011deep}
Glorot, X., Bordes, A., and Bengio, Y. (2011).
\newblock Deep sparse rectifier neural networks.
\newblock In {\em Proceedings of the fourteenth international conference on
  artificial intelligence and statistics}, pages 315--323. JMLR Workshop and
  Conference Proceedings.

\bibitem[Gruntman and Turner, 2013]{gruntman2013integration}
Gruntman, E. and Turner, G.~C. (2013).
\newblock Integration of the olfactory code across dendritic claws of single
  mushroom body neurons.
\newblock {\em Nature neuroscience}, 16(12):1821--1829.

\bibitem[Hall and Kier, 1991]{hall1991molecular}
Hall, L.~H. and Kier, L.~B. (1991).
\newblock The molecular connectivity chi indexes and kappa shape indexes in
  structure-property modeling.
\newblock {\em Reviews in computational chemistry}, pages 367--422.

\bibitem[Honegger et~al., 2011]{honegger2011cellular}
Honegger, K.~S., Campbell, R.~A., and Turner, G.~C. (2011).
\newblock Cellular-resolution population imaging reveals robust sparse coding
  in the drosophila mushroom body.
\newblock {\em Journal of neuroscience}, 31(33):11772--11785.

\bibitem[Huang and Chen, 2007]{huang2007convex}
Huang, G.-B. and Chen, L. (2007).
\newblock Convex incremental extreme learning machine.
\newblock {\em Neurocomputing}, 70(16-18):3056--3062.

\bibitem[Huang et~al., 2006a]{huang2006universal}
Huang, G.-B., Chen, L., Siew, C.~K., et~al. (2006a).
\newblock Universal approximation using incremental constructive feedforward
  networks with random hidden nodes.
\newblock {\em IEEE Trans. Neural Networks}, 17(4):879--892.

\bibitem[Huang et~al., 2006b]{huang2006extreme}
Huang, G.-B., Zhu, Q.-Y., and Siew, C.-K. (2006b).
\newblock Extreme learning machine: theory and applications.
\newblock {\em Neurocomputing}, 70(1-3):489--501.

\bibitem[Igelnik and Pao, 1995]{igelnik1995stochastic}
Igelnik, B. and Pao, Y.-H. (1995).
\newblock Stochastic choice of basis functions in adaptive function
  approximation and the functional-link net.
\newblock {\em IEEE transactions on Neural Networks}, 6(6):1320--1329.

\bibitem[Inada et~al., 2017]{inada2017origins}
Inada, K., Tsuchimoto, Y., and Kazama, H. (2017).
\newblock Origins of cell-type-specific olfactory processing in the drosophila
  mushroom body circuit.
\newblock {\em Neuron}, 95(2):357--367.

\bibitem[Keller and Vosshall, 2016]{keller2016olfactory}
Keller, A. and Vosshall, L.~B. (2016).
\newblock Olfactory perception of chemically diverse molecules.
\newblock {\em BMC neuroscience}, 17(1):1--17.

\bibitem[LeCun et~al., 1998]{lecun1998gradient}
LeCun, Y., Bottou, L., Bengio, Y., and Haffner, P. (1998).
\newblock Gradient-based learning applied to document recognition.
\newblock {\em Proceedings of the IEEE}, 86(11):2278--2324.

\bibitem[Li et~al., 1997]{li1997comments}
Li, J.-Y., Chow, W.~S., Igelnik, B., and Pao, Y.-H. (1997).
\newblock Comments on" stochastic choice of basis functions in adaptive
  function approximation and the functional-link net"[with reply].
\newblock {\em IEEE transactions on neural networks}, 8(2):452--454.

\bibitem[Liang et~al., 2006]{liang2006fast}
Liang, N.-Y., Huang, G.-B., Saratchandran, P., and Sundararajan, N. (2006).
\newblock A fast and accurate online sequential learning algorithm for
  feedforward networks.
\newblock {\em IEEE Transactions on neural networks}, 17(6):1411--1423.

\bibitem[Lin et~al., 2014]{lin2014sparse}
Lin, A.~C., Bygrave, A.~M., De~Calignon, A., Lee, T., and Miesenb{\"o}ck, G.
  (2014).
\newblock Sparse, decorrelated odor coding in the mushroom body enhances
  learned odor discrimination.
\newblock {\em Nature neuroscience}, 17(4):559--568.

\bibitem[Mamlouk and Martinetz, 2004]{mamlouk2004dimensions}
Mamlouk, A.~M. and Martinetz, T. (2004).
\newblock On the dimensions of the olfactory perception space.
\newblock {\em Neurocomputing}, 58:1019--1025.

\bibitem[Meulemans et~al., 2020]{meulemans2020theoretical}
Meulemans, A., Carzaniga, F.~S., Suykens, J.~A., Sacramento, J., and Grewe,
  B.~F. (2020).
\newblock A theoretical framework for target propagation.
\newblock {\em arXiv preprint arXiv:2006.14331}.

\bibitem[Owald and Waddell, 2015]{owald2015olfactory}
Owald, D. and Waddell, S. (2015).
\newblock Olfactory learning skews mushroom body output pathways to steer
  behavioral choice in drosophila.
\newblock {\em Current opinion in neurobiology}, 35:178--184.

\bibitem[Pao and Takefuji, 1992]{pao1992functional}
Pao, Y.-H. and Takefuji, Y. (1992).
\newblock Functional-link net computing: theory, system architecture, and
  functionalities.
\newblock {\em Computer}, 25(5):76--79.

\bibitem[Paszke et~al., 2017]{paszke2017automatic}
Paszke, A., Gross, S., Chintala, S., Chanan, G., Yang, E., DeVito, Z., Lin, Z.,
  Desmaison, A., Antiga, L., and Lerer, A. (2017).
\newblock Automatic differentiation in pytorch.
\newblock In {\em NIPS-W}.

\bibitem[Scardapane and Wang, 2017]{scardapane2017randomness}
Scardapane, S. and Wang, D. (2017).
\newblock Randomness in neural networks: an overview.
\newblock {\em Wiley Interdisciplinary Reviews: Data Mining and Knowledge
  Discovery}, 7(2):e1200.

\bibitem[Schmidt et~al., 1992]{schmidt1992feed}
Schmidt, W.~F., Kraaijveld, M.~A., Duin, R.~P., et~al. (1992).
\newblock Feed forward neural networks with random weights.
\newblock In {\em International Conference on Pattern Recognition}, pages 1--1.
  IEEE COMPUTER SOCIETY PRESS.

\bibitem[Sell, 2006]{sell2006unpredictability}
Sell, C.~S. (2006).
\newblock On the unpredictability of odor.
\newblock {\em Angewandte Chemie International Edition}, 45(38):6254--6261.

\bibitem[Seung, 2012]{seung2012connectome}
Seung, S. (2012).
\newblock {\em Connectome: How the brain's wiring makes us who we are}.
\newblock Houghton Mifflin Harcourt.

\bibitem[Sporns et~al., 2005]{sporns2005human}
Sporns, O., Tononi, G., and K{\"o}tter, R. (2005).
\newblock The human connectome: a structural description of the human brain.
\newblock {\em PLoS Comput Biol}, 1(4):e42.

\bibitem[Takemura et~al., 2017]{takemura2017connectome}
Takemura, S.-y., Aso, Y., Hige, T., Wong, A., Lu, Z., Xu, C.~S., Rivlin, P.~K.,
  Hess, H., Zhao, T., Parag, T., et~al. (2017).
\newblock A connectome of a learning and memory center in the adult drosophila
  brain.
\newblock {\em Elife}, 6:e26975.

\bibitem[van Schaik and Tapson, 2015]{van2015online}
van Schaik, A. and Tapson, J. (2015).
\newblock Online and adaptive pseudoinverse solutions for elm weights.
\newblock {\em Neurocomputing}, 149:233--238.

\bibitem[Vosshall et~al., 1999]{vosshall1999spatial}
Vosshall, L.~B., Amrein, H., Morozov, P.~S., Rzhetsky, A., and Axel, R. (1999).
\newblock A spatial map of olfactory receptor expression in the drosophila
  antenna.
\newblock {\em Cell}, 96(5):725--736.

\bibitem[Wang et~al., 2003]{wang2003two}
Wang, J.~W., Wong, A.~M., Flores, J., Vosshall, L.~B., and Axel, R. (2003).
\newblock Two-photon calcium imaging reveals an odor-evoked map of activity in
  the fly brain.
\newblock {\em Cell}, 112(2):271--282.

\bibitem[Xiao et~al., 2017]{xiao2017fashion}
Xiao, H., Rasul, K., and Vollgraf, R. (2017).
\newblock Fashion-mnist: a novel image dataset for benchmarking machine
  learning algorithms.
\newblock {\em arXiv preprint arXiv:1708.07747}.

\end{thebibliography}

\clearpage

\appendix
\section*{Appendix}
\label{sec:appendix}
\renewcommand\thefigure{A-\arabic{figure}}\setcounter{figure}{0}
\renewcommand\thetable{A-\arabic{table}}
\setcounter{table}{0}
\setcounter{section}{0}
\pagenumbering{Alph}

\section{The Derivative of \texorpdfstring{$\nabla_{s_{ji}}$}{TEXT}}
\label{ap:derivative}
All of our experiments make use of classification tasks, with model outputs generated by applying the softmax function. Consequently, the objective function $\mathcal{L}$ that we use is the categorical cross-entropy loss function. The loss function for the odor-perception task is a binary cross-entropy loss, which is a special form of categorical cross-entropy loss.

Given an input vector $\textbf{x} \in \mathbb{R}^{d}$ and one-hot-encoded target vector $\textbf{y} \in \mathbb{R}^{c}$, consider the case where the output weight matrix $\boldsymbol\beta^{*}$ of the KCNet was selected to best fit the linear system $\boldsymbol\beta^{T}\textbf{h} = \textbf{y}$ where \textbf{h} is the hidden layer output vector of the network, as in:
\begin{gather}
    \textbf{h} = 
    \begin{pmatrix}
        h_{1}(\Hat{\textbf{x}}) \\
        \vdots \\
        h_{\mathcal{B}}(\Hat{\textbf{x}}) 
    \end{pmatrix}_{\mathcal{B} \times 1}
    \; \text{and} \quad
    \boldsymbol\beta^{*} = \begin{pmatrix}
        \boldsymbol\beta_{1}^{T} \\
        \vdots \\
        \boldsymbol\beta_{\mathcal{B}}^{T} \\
    \end{pmatrix}_{\mathcal{B} \times c},
    \label{eq:kcnet_matrices}
\end{gather}
and $\Hat{\textbf{x}}$ is the processed input in Eq.~(\ref{eq:kcnet_intermediate}). Based on this, the prediction $\Hat{\textbf{y}}$ of the KCNet is:
\begin{equation}
    \Hat{\textbf{y}} = \textsc{softmax}(\textbf{z}) = \textsc{softmax}(\boldsymbol\beta^{*^T}\textbf{h}).
\end{equation}
In the backward pass, the batch gradient descent using validation set is performed. Given the validation set, the gradient $\nabla_{s_{ji}}$ can be computed as follows:
\begin{equation}
    \begin{split}
        \nabla s_{ji}
        &\approx \frac{\partial  \mathcal{L}}{\partial w_{ji}} \\
        &= \frac{\partial  \mathcal{L}}{\partial \textbf{z}} \frac{\partial \textbf{z}}{\partial h_{j}} \frac{\partial h_{j}}{\partial \Hat{x}_{j}} \frac{\partial \Hat{x}_{j}}{\partial \Bar{x}_{j}} \frac{\partial \Bar{x}_{j}}{\partial w_{ji}}
    \end{split}
\end{equation}
where
\begin{gather}
    \frac{\partial  \mathcal{L}}{\partial \textbf{z}} = (\Hat{\textbf{y}} - \textbf{y}) \in \mathbb{R}^{c},
    \qquad
    \frac{\partial \textbf{z}}{\partial h_{j}} = 
    \begin{pmatrix}
            \beta_{1j}  \\
            \vdots \\
            \beta_{cj} 
    \end{pmatrix}_{c \times 1},
    \qquad
    \frac{\partial h_{j}}{\partial \Hat{x}_{j}} = 
    \begin{cases}
        1 & h_{j} > 0 \; \text{in the forward pass}, \\
        0 &\text{otherwise}
    \end{cases}
    \\
    \frac{\partial \Hat{x}_{j}}{\partial \Bar{x}_{j}} 
    = 1 - \frac{1}{\mathcal{B}},
    \qquad \text{and} \qquad
    \frac{\partial \Bar{x}_{j}}{\partial w_{ji}} 
    = \begin{cases}
        x_{i} & w_{ji} = 1, \\
        0 &\text{otherwise}
    \end{cases}.
\end{gather}
Because $\partial \Hat{x}_{j} / \partial \Bar{x}_{j}$ is a constant and can be integrated with the learning rate $\alpha$, we can write $\nabla_{s_{ji}}$ concisely as follows:
\begin{equation}
\nabla_{s_{ji}} = 
\begin{cases}
    x_{i} \sum_{k=1}^{c} (\Hat{y}_{k} - y_{k})\beta_{kj}& h_{j} > 0 \; \text{in the forward pass and } \; w_{ji} = 1, \\
    0 &\text{otherwise}
\end{cases}.
\end{equation}

\section{Detailed characteristics of the KCNet}
\label{ap:prop_kcnet}
Although the KCNet is devised from the underlying perspectives of biology and neuroscience, several apparent characteristics of the model can also be interpreted from the perspective of computer science and machine learning.

First of all, determining the weight between the input and the hidden layers in the KCNet can be seen as a special case of random weight generation in a NNRW, and the KCNet also uses a bounded nonlinear piecewise continuous activation function, which is the ReLU in our setting. Therefore, KCNet inherits the \emph{universal approximation} property of NNRWs, which means that a single-hidden-layer network with randomly generated hidden nodes and with a broad type of activation functions can universally approximate any continuous target functions in any compact subset $\mathcal{X} \subseteq \mathbb{R}^{d}$~\citep{huang2006extreme, huang2006universal, huang2007convex}. So, theoretically, the more the number of hidden nodes in the KCNet~(i.e., the wider the hidden layer), the better its performance (see the details of KCNet performance results in Section~\ref{sec:exp}).

The second property of the proposed network is that from the point of view in computer science, the hidden-layer output can be viewed as the output of a hash function. \Citet{dasgupta2017neural} took this perspective to interpret the Kenyon Cells~(KCs) as locality-sensitive tags for odors; the more similar a pair of odors, the more similar their assigned tags. Based on this idea, they developed a brain-inspired class of Locality-Sensitive Hash~(LSH) functions~\citep{andoni2006near} for efficiently searching approximate nearest neighbors of high-dimensional points. Because our formulation is partly derived from those insights, our model also can generate tags that preserve the neighborhood structure of the input points using distance-preserving embedding of points, as a family of LSH. Whereas we have focused on the trade-off between conciseness and expressiveness in the width of the KC hidden layer in artificial neural network applications, \citet{dasgupta2017neural} provided computational theory to better understand the performance of architectures matching the scale of the fly brain. Moreover, in their application, a Hebbian-like update rule was used for unsupervised learning of whether an odor had been experienced before (in contrast to the supervised learning we employed). However, their analysis combined with ours implies that KCNet may be employed in unsupervised learning applications, and there will likely be a similar trade-off between the width of the hidden layer and the overall performance of the trained networks.

Finally, to mimic the KCs receiving global inhibition from each other, we subtract the mean of all hidden nodes outputs from each intermediate output in Eq.~(\ref{eq:kcnet_intermediate}). That computation can be viewed as a form of the \emph{layer normalization} in machine learning~\citep{ba2016layer}, which helps to maintain the contributions of each feature so that the network can be unbiased, reduce the internal covariance shift, and accelerate the optimization process. Thus, our model helps to import insights from machine learning into the interpretation of the adaptive function of global inhibition in neuronal structures like the mushroom body.

\section{Complete algorithm of the Dynamic Optimization Algorithm (DOA)}
\begin{algorithm}[ht!]
   \caption{Dynamic Optimization Algorithm (DOA) for Feature Selection}
   \label{alg:doa}
\begin{algorithmic}[1]
   \STATE {\bfseries Input:} $\mathcal{D} = \{\textbf{x}_i \in \mathbb{R}^{d}, \textbf{y}_i \in \mathbb{R}^{c} \}, i=1,\dots,N$ (Dataset, excluding testing set, with the size $N$ samples),
   \STATE {\bfseries Input:} $\mathcal{B} \in \mathbb{R}$ (The number of hidden units),
   \STATE {\bfseries Input:} $\alpha \in \mathbb{R}$ (Learning rate)
   \STATE {\bfseries Output:} $\textbf{W} \in \mathbb{R}^{\mathcal{B} \times d}$ (The weight matrix between the input and the hidden layers),
   \STATE {\bfseries Output:} $\boldsymbol\beta \in \mathbb{R}^{c \times \mathcal{B}}$ (The weight matrix between the hidden and the output layers) \\
   \algorithmiccomment{/*Initialization*/} \\
   \STATE Create the initial weight matrix $\textbf{W} \in \mathcal{R}^{\mathcal{B} \times d}$ and the score matrix $\textbf{S} \in \mathcal{R}^{\mathcal{B} \times d}$
   \REPEAT
   \STATE Randomly split $\mathcal{D}$ into training set $\mathcal{D}_{train}$ and validation set $\mathcal{D}_{val}$ \\
   \algorithmiccomment{/*In forward pass*/} \\
   \STATE Build a model $\mathcal{M}$ using \textbf{W} and train $\boldsymbol\beta$ using the training set $\mathcal{D}_{train}$ 
   \STATE $\mathcal{P} \leftarrow$ Evaluate the model $\mathcal{M}$ using validation set $\mathcal{D}_{val}$ 
   
   \IF{\emph{stopping criteria} are satisfied}
   \STATE break
   \ENDIF 
   
   
   \algorithmiccomment{/*In backward pass*/} \\
   \STATE Compute $\nabla_{\textbf{S}}$ using the validation set $\mathcal{D}_{val}$
   \STATE $\textbf{S} \leftarrow \textbf{S} - \alpha \nabla_{\textbf{S}}$ 
   \STATE $\textbf{S} \leftarrow \text{clip}(\textbf{S}, -1, 1)$
   \STATE $\textbf{W} \leftarrow \rho_{\textbf{S} \rightarrow \textbf{W}}(\textbf{S})$
   \UNTIL{}
\end{algorithmic}
\end{algorithm}

\section{Details of Experiments}
\label{ap:exp}

In this section, we describe the details of our experimental results. The following details apply in common to the two tasks we addressed, odor perception and image classification. Details specific to each task are described in subsections of each task. All the following results were averaged over five experimental replications.

\subsection{Hardware specification of the server}
The hardware specification of the server that we used to experiment is as follows:
\begin{itemize}
    \item CPU: Intel\textregistered{} Core\textsuperscript{TM} i7-6950X CPU @ 3.00GHz (up to 3.50 GHz)
    \item RAM: 128 GB (DDR4 2400MHz)
    \item GPU: NVIDIA GeForce Titan Xp GP102 (Pascal architecture, 3840 CUDA Cores @ 1.6 GHz, 384 bit bus width, 12 GB GDDR G5X memory)
\end{itemize}

\subsection{Source code}
All source codes are available at \url{https://github.com/PavlicLab/KCNet-2021-Hong}. They have been provided as electronic supplementary material as well.

\subsection{Hyperparameter search space}
Table~\ref{tab:hyperparam_search_space} shows the hyperparameter search space of all models for all tasks. To optimize over this hyperparameter space, we used the Optuna~\citep{akiba2019optuna} framework with 100 iterations. It is possible to dynamically construct the parameter search space because the framework allows users to formulate hyperparameter optimization as the maximization/minimization process of an objective function that takes a set of hyperparameters as input and returns a validation score. Furthermore, it provides efficient sampling methods, such as relational sampling that exploits the correlations among the parameters. 

The hyperparameter search space for other non-KCNet machine learning algorithms used as benchmarks were taken to match \citet{chacko2020data} for odor perception tasks. 


To set the number of submodels for the Ensemble-DOA-applied KCNet, we empirically determined that it was the most efficient to divide a large-sized weight matrix of the KCNet into at least 10 small matrices for the tasks that we addressed. Furthermore, we set a value such that the product of the number of submodels and the number of hidden nodes for each submodel in the Ensemble-DOA-applied KCNet equals the number of hidden nodes in the KCNet or DOA-applied KCNet. By doing so, they have the exact dimension of the output weight matrix.

\begin{table}[ht!]
  \caption{Hyperparameter Search Space}
  \centering
  \begin{tabular}{lll}
    \toprule
    Model     & Parameters     & Search Space \\
    \midrule
    \multirow{3}{*}{Adaboost} & \multicolumn{1}{l}{learning\_rate} & \multicolumn{1}{l}{[0.1, 1.0]} \\
                              & \multicolumn{1}{l}{n\_estimators} & \multicolumn{1}{l}{[10, 100]} \\
                              & \multicolumn{1}{l}{algorithm} & \multicolumn{1}{l}{\{'SAMME', 'SAMME.R'\}} \\
    \hline
    \multirow{7}{*}{Gradient Boosting Machine (GBM)} & \multicolumn{1}{l}{learning\_rate} & \multicolumn{1}{l}{[0.05, 1.0]} \\
                              & \multicolumn{1}{l}{n\_estimators} & \multicolumn{1}{l}{[10, 100]} \\
                              & \multicolumn{1}{l}{max\_depth} & \multicolumn{1}{l}{[2, 5]} \\
                              & \multicolumn{1}{l}{min\_samples\_split} & \multicolumn{1}{l}{[10, 200]} \\
                              & \multicolumn{1}{l}{min\_samples\_leaf} & \multicolumn{1}{l}{[30, 70]} \\
                              & \multicolumn{1}{l}{max\_features} & \multicolumn{1}{l}{[8, 159]} \\
                              & \multicolumn{1}{l}{subsample} & \multicolumn{1}{l}{[0.6, 1.0]} \\
    \hline
    \multirow{9}{*}{XGBoost} & \multicolumn{1}{l}{max\_depth} & \multicolumn{1}{l}{[2, 5]} \\
                              & \multicolumn{1}{l}{min\_child\_weight} & \multicolumn{1}{l}{[1, 6]} \\
                              & \multicolumn{1}{l}{gamma} & \multicolumn{1}{l}{[0.1, 10]} \\
                              & \multicolumn{1}{l}{subsample} & \multicolumn{1}{l}{[0.6, 1]} \\
                              & \multicolumn{1}{l}{colsample\_bytree} & \multicolumn{1}{l}{[0.5, 1]} \\
                              & \multicolumn{1}{l}{reg\_alpha} & \multicolumn{1}{l}{[0.1, 20]} \\
                              & \multicolumn{1}{l}{reg\_lambda} & \multicolumn{1}{l}{[0.001, 100]} \\
                              & \multicolumn{1}{l}{learning\_rate} & \multicolumn{1}{l}{[0.01, 1.0]} \\
                              & \multicolumn{1}{l}{n\_estimators} & \multicolumn{1}{l}{[10, 200]} \\
    \hline
    \multirow{5}{*}{Random Forest} & \multicolumn{1}{l}{criterion} & \multicolumn{1}{l}{\{'gini', 'entropy'\}} \\
                              & \multicolumn{1}{l}{n\_estimators} & \multicolumn{1}{l}{[50, 100]} \\
                              & \multicolumn{1}{l}{max\_depth} & \multicolumn{1}{l}{[2, 5]} \\
                              & \multicolumn{1}{l}{min\_samples\_split} & \multicolumn{1}{l}{[2, 70]} \\
                              & \multicolumn{1}{l}{min\_samples\_leaf} & \multicolumn{1}{l}{[1, 50]} \\
    \hline
    \multirow{3}{*}{K-Nearest Neighbors (KNN)} & \multicolumn{1}{l}{weights} & \multicolumn{1}{l}{\{'uniform', 'distance'\}} \\
                              & \multicolumn{1}{l}{p} & \multicolumn{1}{l}{\{1, 2, 3\}} \\
                              & \multicolumn{1}{l}{n\_neighbors} & \multicolumn{1}{l}{[2, 30]} \\
    \hline
    Support Vector Machine (SVM) & C & [2e-3, 2e7] \\
    \hline
    KCNet &  lambda & [2e-2, 2e3] \\
    \hline
    \multirow{3}{*}{KCNet w/ DOA} & \multicolumn{1}{l}{n\_epochs} & \multicolumn{1}{l}{\{5, 10, 50, 100\}} \\
                              & \multicolumn{1}{l}{learning\_rate} & \multicolumn{1}{l}{[5e-1, 5e-7]} \\
                              & \multicolumn{1}{l}{stop\_metric} & \multicolumn{1}{l}{[0.80, 0.97]} \\
    \hline
    \multirow{3}{*}{KCNet w/ Ensemble DOA} & \multicolumn{1}{l}{n\_epochs} & \multicolumn{1}{l}{\{5, 10, 50, 100\}} \\
                              & \multicolumn{1}{l}{learning\_rate} & \multicolumn{1}{l}{[5e-1, 5e-7]} \\
                              & \multicolumn{1}{l}{stop\_metric} & \multicolumn{1}{l}{[0.80, 0.97]} \\
    \bottomrule
  \end{tabular}
  \label{tab:hyperparam_search_space}
\end{table}

\subsection{Odor Perception}

\subsubsection{Methods and Implementations} 
\begin{figure}[t!]
\vskip 0.2in
\begin{center}
\centerline{\includegraphics[width=\columnwidth]{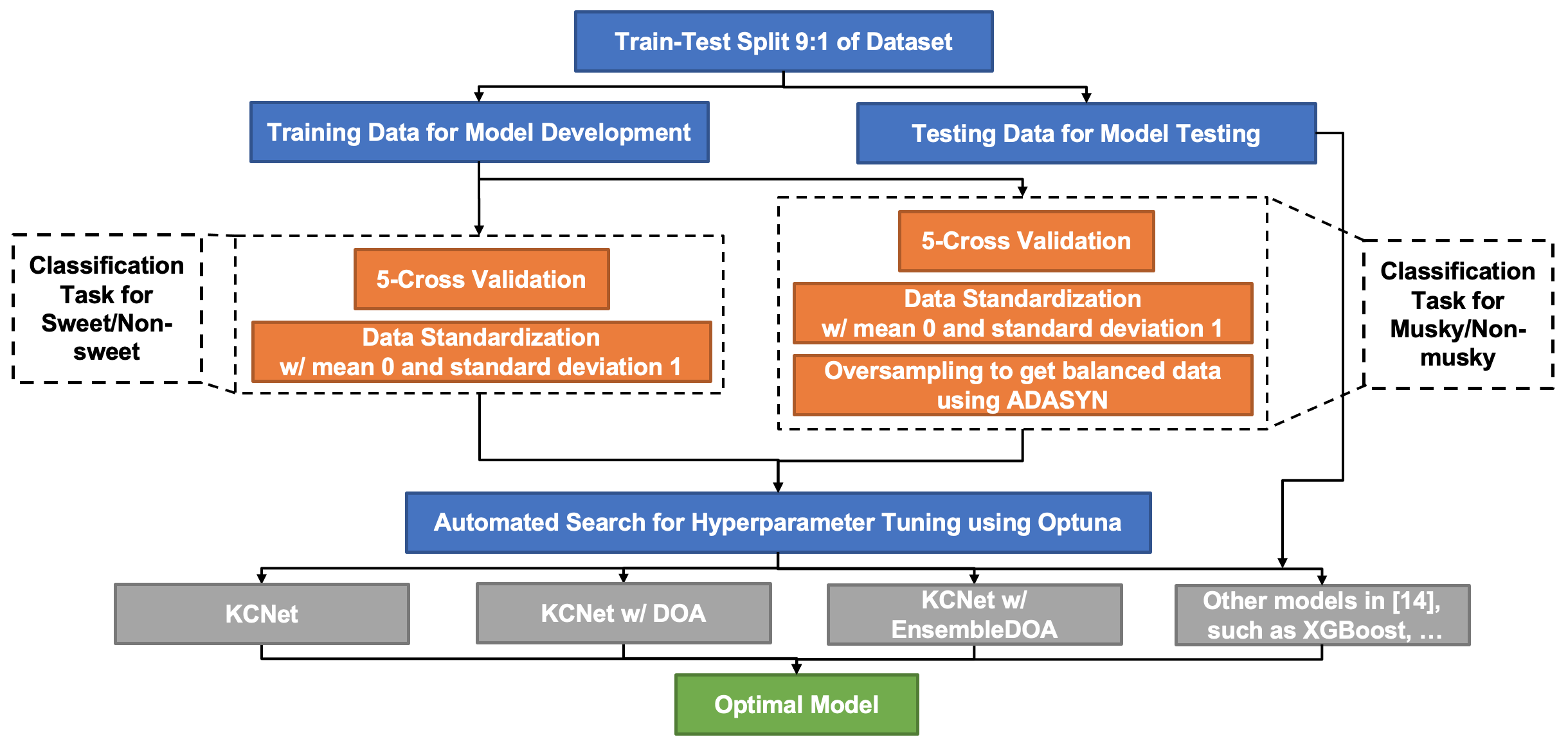}}
\caption{Overall workflow of model development for odor perception tasks}
\label{fig:odor_perception_workflow}
\end{center}
\vskip -0.2in
\end{figure}

Figure~\ref{fig:odor_perception_workflow} depicts the overall workflow for odor-perception tasks. We used the compound identification~(CID) number of 480 odorants included in the original dataset to generate the SMILES notation of the molecules, which was utilized to compute the RDKit descriptors. We obtained the SMILES notations of the odorant compounds using the PubChemPy package of Python programming language. RDKit descriptors were calculated using ChemDes, a free web-based platform for the calculation of molecular descriptors. All the preprocessing steps and model training was performed using Python libraries: NumPy, pandas, imblearn, and scikit-learn. All models that we tested are provided by scikit-learn, and we also implemented our proposed model compatible with scikit-learn. The ADASYN class of the imblearn package was used to scale and oversample the training data during the five-fold cross-validation. In addition, we set the \texttt{random\_state} to 1 for reproducing our results readily. Figures are generated using Matplotlib and Seaborn libraries supported by Python programming language.

\subsubsection{Results}
\begin{figure*}
\centering
\begin{subfigure}[b]{0.45\textwidth}
 \centering
 \includegraphics[width=\textwidth]{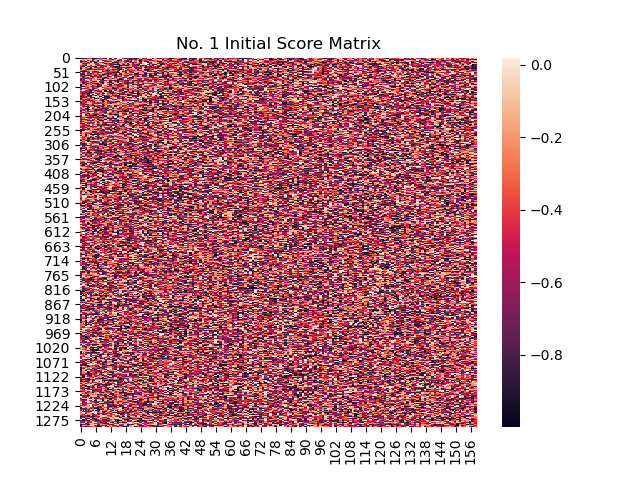}
 \caption{Initial preference score matrix of the KCNet with 1,300 hidden nodes (x-axis: dim. of input features, y-axis: dim. of hidden units)}
 \label{fig:doa_sweet_init_s}
\end{subfigure}
\hfill
\begin{subfigure}[b]{0.45\textwidth}
 \centering
 \includegraphics[width=\textwidth]{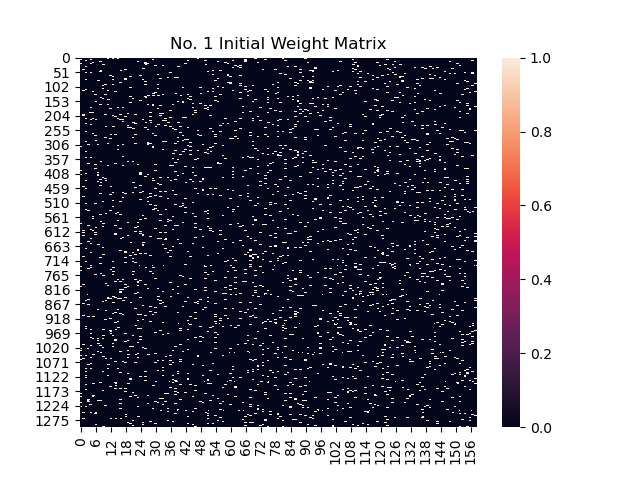}
 \caption{Initial weight matrix of the KCNet with 1,300 hidden nodes (x-axis: dim. of input features, y-axis: dim. of hidden units)}
 \label{fig:doa_sweet_init_w}
\end{subfigure}
\hfill
\begin{subfigure}[b]{0.45\textwidth}
 \centering
 \includegraphics[width=\textwidth]{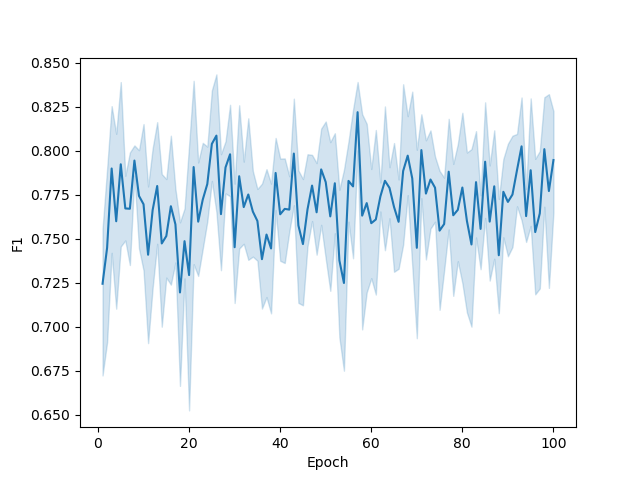}
 \caption{Validation weighted F1 score during the DOA with 100 epochs. (x-axis: \# of epochs, y-axis: validation weighted F1 score)}
 \label{fig:doa_sweet_val_f1}
\end{subfigure}
\hfill
\begin{subfigure}[b]{0.45\textwidth}
 \centering
 \includegraphics[width=\textwidth]{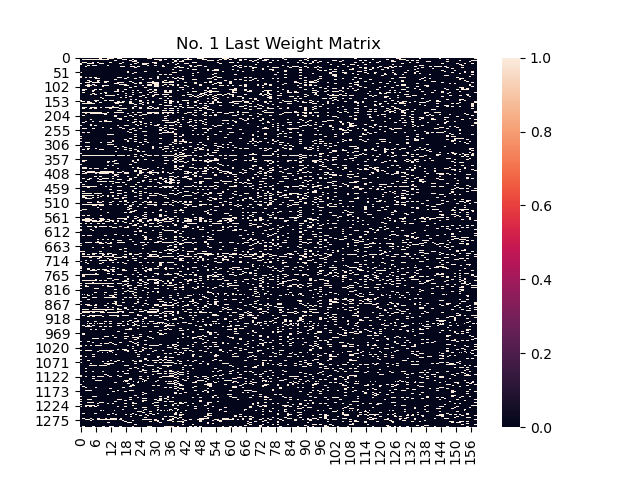}
 \caption{Final weight matrix of the KCNet with 1300 hidden nodes after the DOA (x-axis: dim. of input features, y-axis: dim. of hidden units)}
 \label{fig:doa_sweet_last_w}
\end{subfigure}
\hfill
\begin{subfigure}[b]{0.45\textwidth}
 \centering
 \includegraphics[width=\textwidth]{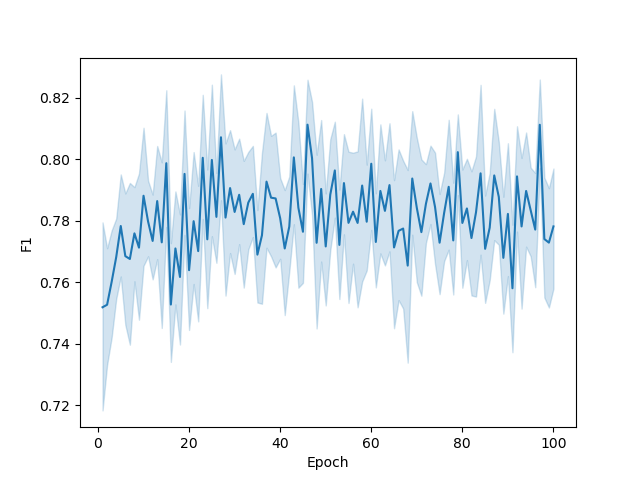}
 \caption{Validation weighted F1 score of 13 submodels during the Ensemble DOA with 100 epochs x-axis: \# of epochs, y-axis: validation weighted F1 score)}
 \label{fig:sweet_ensemble_doa_val_f1}
\end{subfigure}
\hfill
\begin{subfigure}[b]{0.45\textwidth}
 \centering
 \includegraphics[width=\textwidth]{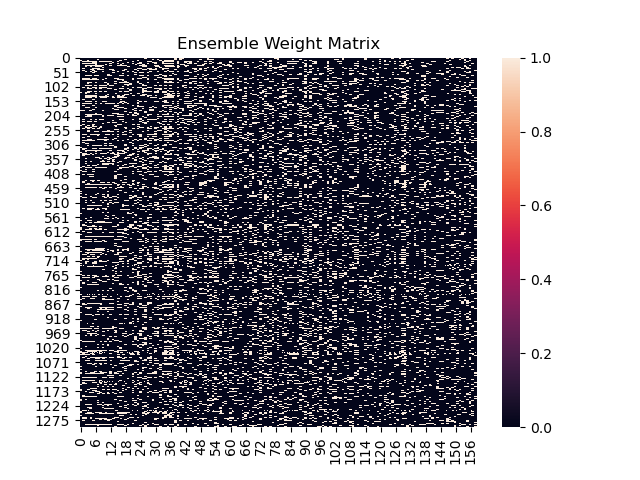}
 \caption{Final weight matrix of the Ensemble-DOA-KCNet concatenated by 13 submodels with 100 hidden nodes for each submodel (x-axis: dim. of input features, y-axis: dim. of hidden units)}
 \label{fig:sweet_ensemble_doa_last_w}
\end{subfigure}
\hfill
\caption{Results of the KCNet, the DOA-applied KCNet, and the Ensemble-DOA-applied KCNet for Sweet/Non-sweet}
\label{fig:results_sweet}
\end{figure*}

Figure~\ref{fig:results_sweet} shows the results about the KCNet, the DOA-applied KCNet, and the Ensemble-DOA-applied KCNet for Sweet/Non-sweet OC. Figures~\ref{fig:doa_sweet_init_s} and~\ref{fig:doa_sweet_init_w} represent the initial preference score and weight matrix, respectively, before applying the DOA to the KCNet. Figure~\ref{fig:doa_sweet_val_f1} depicts the change in the validation-weighted F1 score during the DOA, and this resulted in the final weight matrix of the KCNet shown in Fig.~\ref{fig:doa_sweet_last_w}. Compared to the initial weight matrix in Fig.~\ref{fig:doa_sweet_init_w}, the final matrix was more densified to enable the exploitation of the combination of inputs. We applied the Ensemble DOA to the KCNet by dividing the weight matrix with 1,300 hidden nodes into 13 submodels with 100 hidden nodes for each submodel. Figure~\ref{fig:sweet_ensemble_doa_val_f1} demonstrates the change of the validation-weighted F1 score of 13 submodels during the Ensemble DOA and the final weight matrix assembled by 13 submodels was shown in Fig.~\ref{fig:sweet_ensemble_doa_last_w}. Compared to the weight matrix of the DOA-applied KCNet in Fig.~\ref{fig:doa_sweet_last_w}, it appears to have a slightly more regular pattern, which seems to be due to the concatenation of multiple small matrices. The best hyperparameter settings for the KCNet, the DOA-applied KCNet, and the Ensemble-DOA-applied KCNet for \emph{Sweet} OC as follows:
\begin{itemize}
    \item The KCNet : \texttt{lambda} = 100
    \item The DOA-applied KCNet : \texttt{n\_epoch} = 100, \texttt{learning\_rate} = 0.5, \texttt{stop\_metric} = 0.87
    \item The Ensemble-DOA-applied KCNet : \texttt{n\_epoch} = 100, \texttt{learning\_rate} = 0.5, \texttt{stop\_metric} = 0.90
\end{itemize}

\begin{figure*}
\centering
\begin{subfigure}[b]{0.45\textwidth}
 \centering
 \includegraphics[width=\textwidth]{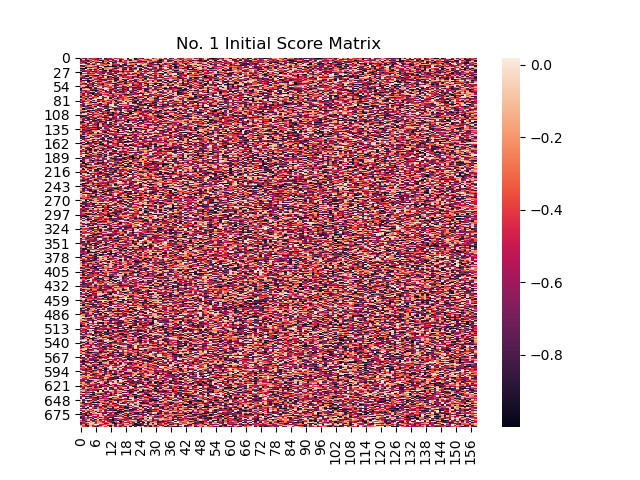}
 \caption{Initial preference score matrix of the KCNet with 700 hidden nodes (x-asis: dim. of input features, y-axis: dim. of hidden units)}
 \label{fig:doa_musky_init_s}
\end{subfigure}
\hfill
\begin{subfigure}[b]{0.45\textwidth}
 \centering
 \includegraphics[width=\textwidth]{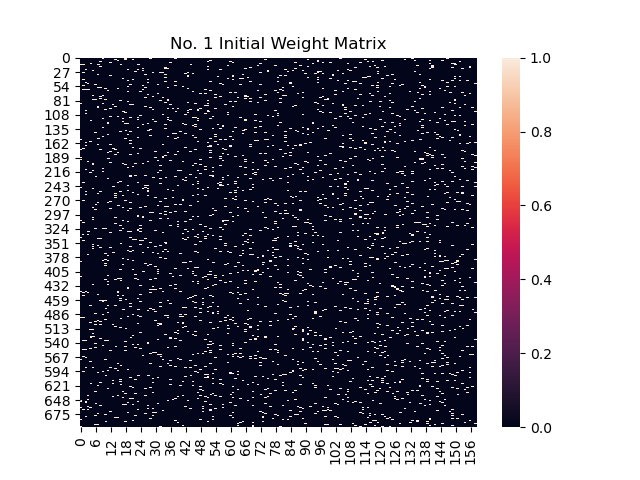}
 \caption{Initial weight matrix of the KCNet with 700 hidden nodes (x-asis: dim. of input features, y-axis: dim. of hidden units)}
 \label{fig:doa_musky_init_w}
\end{subfigure}
\hfill
\begin{subfigure}[b]{0.45\textwidth}
 \centering
 \includegraphics[width=\textwidth]{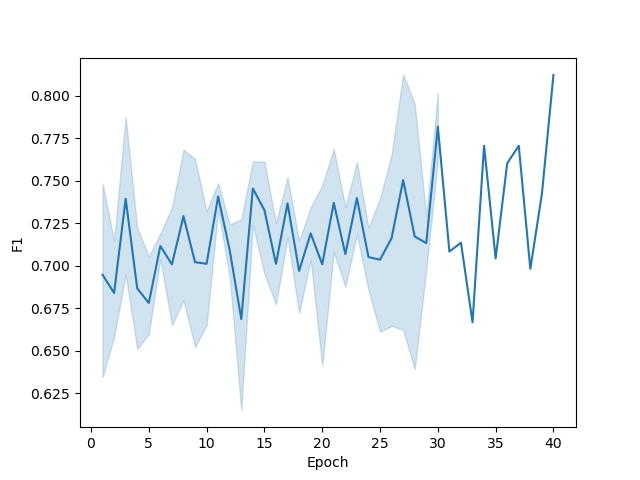}
 \caption{Validation weighted F1 score during the DOA with 100 epochs (x-axis: \# of epochs, y-axis: validation weighted F1 score)}
 \label{fig:doa_musky_val_f1}
\end{subfigure}
\hfill
\begin{subfigure}[b]{0.45\textwidth}
 \centering
 \includegraphics[width=\textwidth]{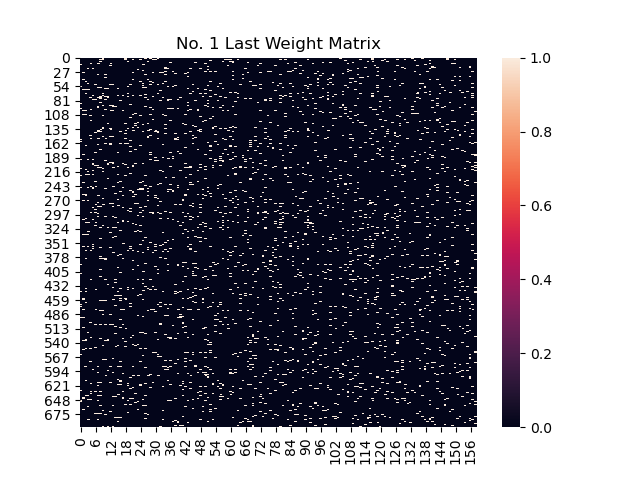}
 \caption{Final weight matrix of the KCNet with 700 hidden nodes after the DOA (x-asis: dim. of input features, y-axis: dim. of hidden units)}
 \label{fig:doa_musky_last_w}
\end{subfigure}
\hfill
\begin{subfigure}[b]{0.45\textwidth}
 \centering
 \includegraphics[width=\textwidth]{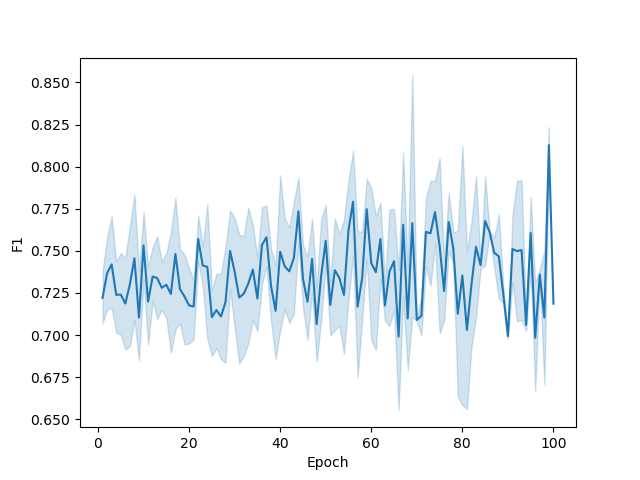}
 \caption{Validation weighted F1 score of 10 submodels during Ensemble DOA with 100 epochs (x-axis: \# of epochs, y-axis: validation weighted F1 score)}
 \label{fig:musky_ensemble_doa_val_f1}
\end{subfigure}
\hfill
\begin{subfigure}[b]{0.45\textwidth}
 \centering
 \includegraphics[width=\textwidth]{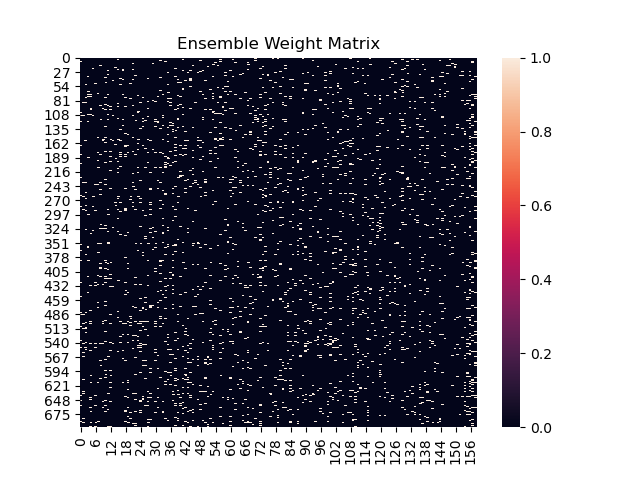}
 \caption{Final weight matrix of the Ensemble-DOA-applied KCNet concatenated by 10 submodels with 70 hidden nodes for each submodel (x-asis: dim. of input features, y-axis: dim. of hidden units)}
 \label{fig:musky_ensemble_doa_last_w}
\end{subfigure}
\hfill
\caption{Results of the KCNet the DOA-applied KCNet, and the Ensemble-DOA-applied KCNet for Musky/Non-musky}
\label{fig:results_musky}
\end{figure*}

Similar to the above description, Fig.~\ref{fig:results_musky} demonstrates the results about the KCNet, the DOA-applied KCNet, and the Ensemble-DOA-applied KCNet for Musky/Non-musky OC. The best hyperparameter settings for the KCNet, the DOA-applied KCNet and the Ensemble-DOA-applied KCNet for \emph{Musky} OC as follows:
\begin{itemize}
    \item The KCNet : \texttt{lambda} = 115
    \item The DOA-applied KCNet : \texttt{n\_epoch} = 100, \texttt{learning\_rate} = 0.5, \texttt{stop\_metric} = 0.80
    \item The Ensemble-DOA-applied KCNet : \texttt{n\_epoch} = 100, \texttt{learning\_rate} = 0.1, \texttt{stop\_metric} = 0.82
\end{itemize}

\subsection{Image Classification}

\subsubsection{Methods and implementations}
Here, we explain the details of implementing the Fully-trained Single-Hidden-layer Neural Network~(FSHN). Compared to the KCNet, all the parameters, including hidden weights in the FSHN, are adaptable. Thus, for the sake of accurate comparison of performance, it is important to set the number of learnable parameters of the FSHN as similar to those of the KCNet. The number of adaptable parameters for the KCNet is determined by multiplying the number of hidden nodes by the number of output nodes. So, the numbers of learnable parameters of the KCNet for MNIST, Fashion-MNIST, and EMNIST-Balanced are 65,000 (6,500 $\times$ 10), 70,000 (7,000 $\times$ 10), and 211,500 (4,500 $\times$ 47), respectively, whereas the numbers of adaptable parameters of the FSHN are 65200, 70765, and 211,375, respectively, as shown in Table~\ref{tab:result_image}. Because the FSHN exploits a usual linear model supported by PyTorch and it includes biases as parameters, we set them to match the number of parameters of the KCNet as closely as possible. The number of hidden nodes in the FSHN that we set can be seen in the source codes that we provided. 

For the hyperparameter setting for the FSHN, by referring to the basic tutorial of constructing a classifier using MNIST that PyTorch provides, we define the FSHN as follows: \texttt{training\_batch\_size} = 64, \texttt{test\_batch\_size} = 1000, \texttt{epoch} = 5, \texttt{learning\_rate} = 0.1. This setting was shared for other image classification tasks. The reason why the number of epochs was set to 5 for the FSHN was the number that allows the FSHN to achieve better performance for MNIST than the KCNet. 

\subsubsection{Analysis of the DOA}
In this section, we explain the limitation of the DOA and how the limit resulted in the invention of the Ensemble DOA. To analyze the effectiveness of applying the DOA to the KCNet, we tested our model and optimization method using the MNIST dataset. It was not appropriate to analyze the performance of the DOA using the odor-perception datasets because, as we mentioned, the number of data samples for odor perception was not enough (only 960 data samples), and the dataset had the inherited characteristic of imbalance. In contrast, the MNIST dataset is large enough and has a balanced distribution of all classes. 

\begin{figure*}[t!]
\centering
\begin{subfigure}[b]{0.4\textwidth}
 \centering
 \includegraphics[width=\textwidth]{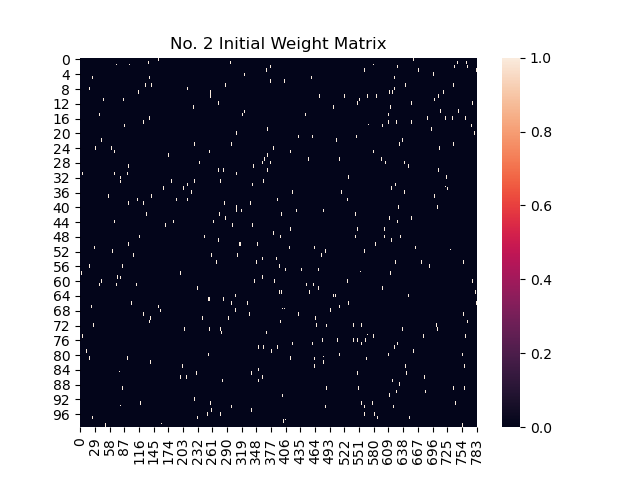}
 \caption{Initial weight matrix with 100 hidden nodes before the DOA}
 \label{fig:weight_before_doa}
\end{subfigure}
\hfill
\begin{subfigure}[b]{0.4\textwidth}
 \centering
 \includegraphics[width=\textwidth]{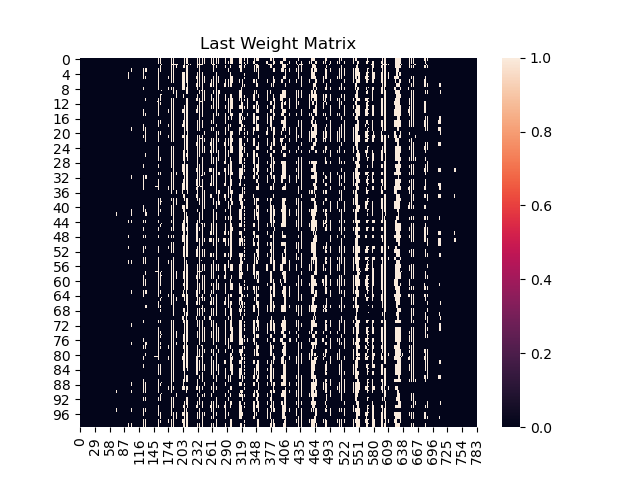}
 \caption{Final weight matrix with 100 hidden nodes after the DOA with 50 epochs}
 \label{fig:weight_after_doa}
\end{subfigure}
\hfill
\begin{subfigure}[b]{0.38\textwidth}
 \centering
 \includegraphics[width=\textwidth]{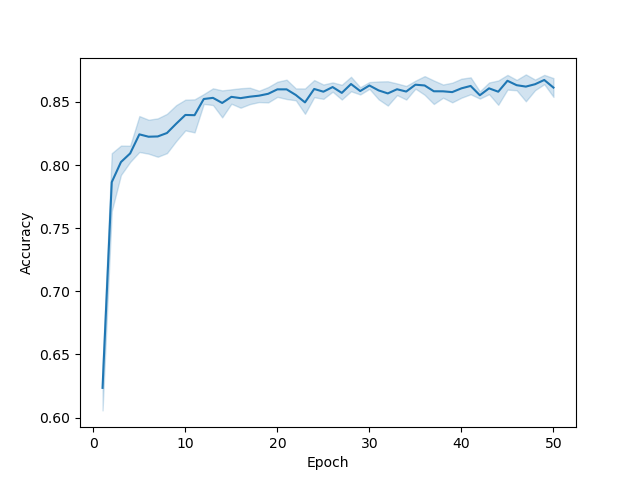}
 \caption{Validation accuracy with 100 hidden nodes during the DOA with 50 epochs}
 \label{fig:doa_hsize-100}
\end{subfigure}
\hfill
\begin{subfigure}[b]{0.38\textwidth}
 \centering
 \includegraphics[width=\textwidth]{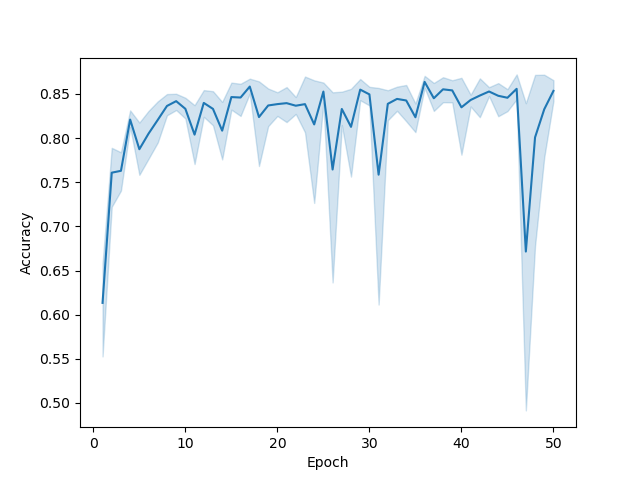}
 \caption{Validation accuracy with 500 hidden nodes during the DOA with 50 epochs}
 \label{fig:doa_hsize-500}
\end{subfigure}
\hfill
\begin{subfigure}[b]{0.38\textwidth}
 \centering
 \includegraphics[width=\textwidth]{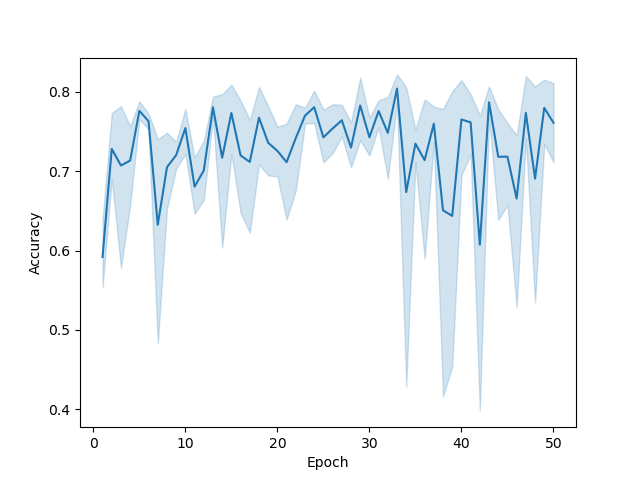}
 \caption{Validation accuracy with 1,000 hidden nodes during the DOA with 50 epochs}
 \label{fig:doa_hsize-1000}
\end{subfigure}
\hfill
\begin{subfigure}[b]{0.38\textwidth}
 \centering
 \includegraphics[width=\textwidth]{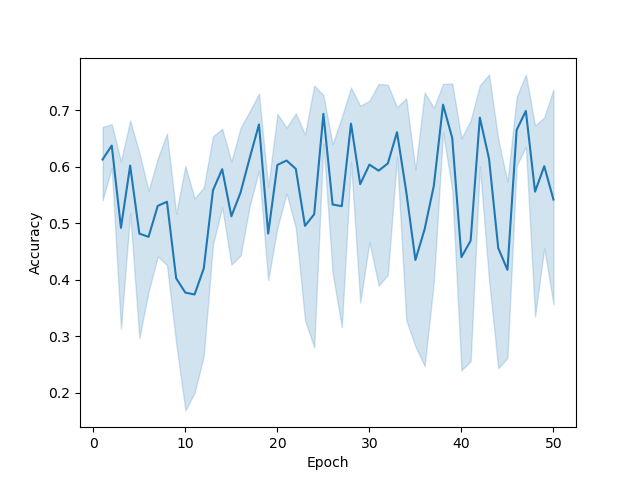}
 \caption{Validation accuracy with 2,000 hidden nodes during the DOA with 50 epochs}
 \label{fig:doa_hsize-2000}
\end{subfigure}
\hfill
\caption{For image classification using MNIST, the change of weight matrix of the KCNet before and after the DOA and the relationship between validation accuracy depending on its hidden size and learning rate = 0.05}
\label{fig:relation_hsize_lr}
\end{figure*}

Figure~\ref{fig:relation_hsize_lr} depicts the efficiency of applying the DOA to the KCNet with a small weight matrix and the increased sensitivity of the KCNet to the learning rate as the number of hidden units increases. Figures~\ref{fig:weight_before_doa} and~\ref{fig:weight_after_doa} show a noticeable change in the weight matrix of the KCNet before and after the DOA. This indicates the intuitive effectiveness of applying the DOA to the KCNet. The method allows the model to explore a useful combination of inputs, and the majority of meaningful features in MNIST images were chosen and exploited to allow the model to have a powerful discrimination ability. However, this phenomenon can likely be shown when applying the DOA to the KCNet with a relatively small weight matrix. 

\begin{figure*}[t!]
\centering
\begin{subfigure}[b]{0.48\textwidth}
 \centering
 \includegraphics[width=\textwidth]{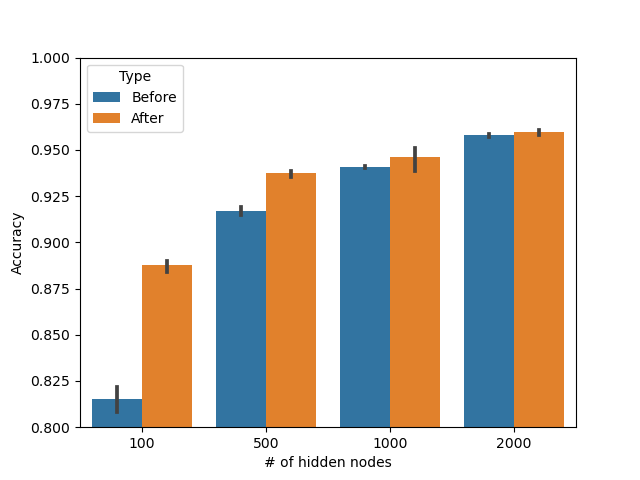}
 \caption{The change of test accuracy before and after applying the DOA}
 \label{fig:result_before_after_doa}
\end{subfigure}
\hfill
\begin{subfigure}[b]{0.48\textwidth}
 \centering
 \includegraphics[width=\textwidth]{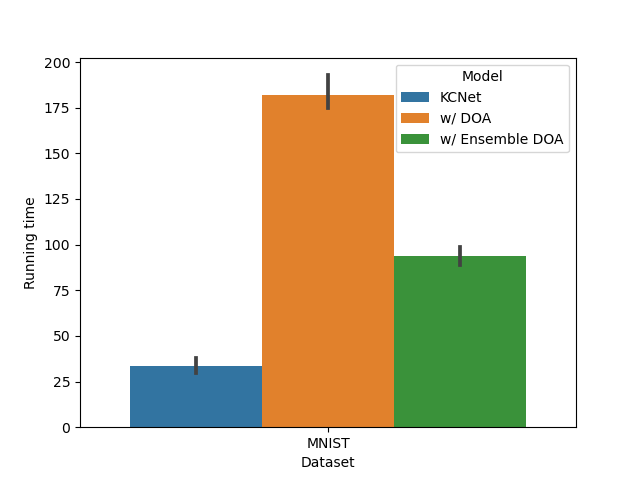}
 \caption{The running time (sec) for the KCNet variants}
 \label{fig:running_time_kcnet_variants_mnist}
\end{subfigure}
\hfill
\caption{For image classification using MNIST, the change of test accuracy of the KCNet before and after the DOA and the running time comparison.}
\label{fig:perfomance_analysis}
\end{figure*}

From Figs.~\ref{fig:doa_hsize-100} to~\ref{fig:doa_hsize-2000}, the convergence of validation accuracy of the KCNet with the same value of learning rate, which is 0.5, during the DOA was shown as the number of its hidden nodes increases. As shown in Fig.~\ref{fig:doa_hsize-100}, the validation of the accuracy of the KCNet with 100 hidden nodes was converged well, and this resulted in a notable transformation of the weight matrix shown in Figs.~\ref{fig:weight_before_doa} and~\ref{fig:weight_after_doa}. This corresponded to the substantial increase in test accuracy in Fig.~\ref{fig:result_before_after_doa}. However, the larger the weight matrix of the KCNet, the more perturbed the convergence of validation accuracy and the more sensitive the KCNet to the learning rate. This led to a marginal improvement of the DOA-applied KCNet as its weight matrix is too large to properly deal with the complexity of exploring a set of inputs in Fig.~\ref{fig:result_before_after_doa}. Furthermore, the larger the weight matrix, the more burdensome the computation of the backward pass of the DOA. 

Thus, to reduce the computational burden of the backward pass and to exploit the full power of the DOA, we devised the Ensemble DOA, the ensemble method of assembling a weight matrix by concatenating several smaller matrices after applying the DOA to each small matrix. Figure~\ref{fig:running_time_kcnet_variants_mnist} demonstrates the running time, including the model configuration time, training time, and evaluation time, of the KCNet, the DOA-applied KCNet. The running time refers to the time spent running on a single CPU and RAM because we did not want to include any redundant time, such as the time to exchange the data between CPU and GPU. As shown in Fig.~\ref{fig:running_time_kcnet_variants_mnist}, the running time of the Ensemble-DOA-applied KCNet is much faster than the DOA-applied KCNet, and the application of the Ensemble DOA allows the KCNet to achieve better performance than the DOA-applied KCNet. The results can be shown in Tables~\ref{tab:result_odor} and~\ref{tab:result_image} for all tasks.


\subsubsection{Results}
\begin{figure*}\centering
    \hfill
    \begin{subfigure}[b]{0.48\textwidth}\centering
        \includegraphics[width=\textwidth]{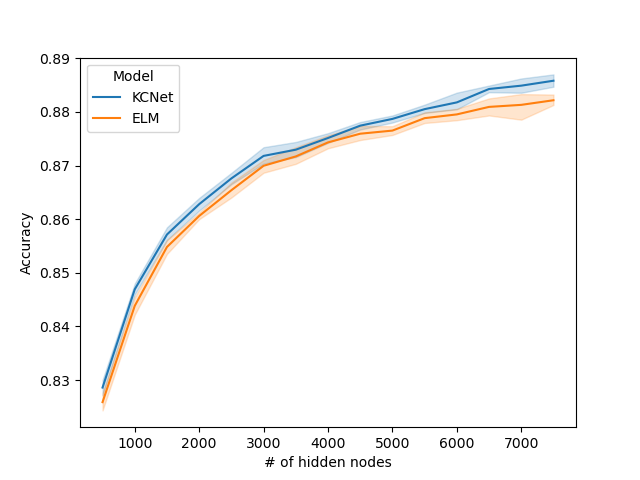}
        \caption{Test accuracy of KCNet and ELM for Fashion-MNIST depending on their hidden sizes}
        \label{fig:fashioin_mnist_kcnet_elm_hsize}
    \end{subfigure}
    \hfill
    \begin{subfigure}[b]{0.48\textwidth}\centering
        \includegraphics[width=\textwidth]{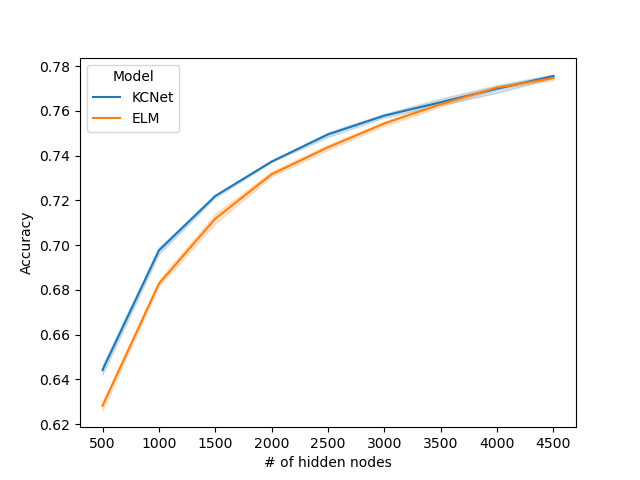}
        \caption{Test accuracy of KCNet and ELM for EMNIST-Balanced depending on their hidden sizes}
        \label{fig:emnist_kcnet_elm_hsize}
    \end{subfigure}
    \hfill
    \caption{Test accuracy of KCNet and ELM for Fashion-MNIST and EMNIST-Balanced depending on their hidden sizes}
    \label{ap:fashion_mnist_emnist_kcnet_elm_hsize}
\end{figure*}

Figure~\ref{fig:running_time_kcnet_elm_fshn} in the main text shows the running time, including model configuration, training, and evaluation time of the KCNet, ELM, and FSHN using a single CPU and RAM only (i.e., not making use of a GPU). This is because we wanted to exclude any redundant time such as the data transmission time between GPU and GPU.

Figure~\ref{ap:fashion_mnist_emnist_kcnet_elm_hsize} (and~\ref{fig:mnist_kcnet_elm_hsize} in the main text) shows that the KCNet mostly outperforms the ELM on all image benchmark datasets, depending on their size of hidden nodes. We set the following range of hidden nodes to get the result within the limitation of RAM of GPU we used; $\{500, 1000, \dots, 6500\}$ for MNIST, $\{500, 1000, \dots, 7500\}$ for Fashion-MNIST, and $\{500, 1000, \dots, 4500\}$ for EMNIST-Balanced. 

\begin{figure*}[t!]
\centering
    \hfill
    \begin{subfigure}[b]{0.48\textwidth}\centering
        \includegraphics[width=\textwidth]{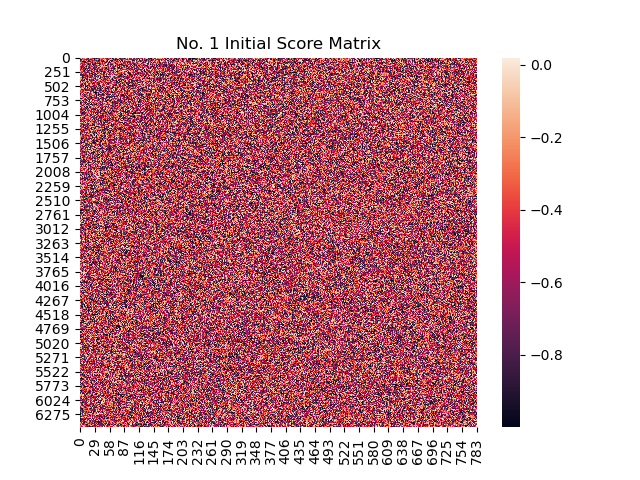}
        \caption{Initial preference score matrix of the KCNet with 6,500 hidden nodes before the DOA (x-axis: dim. of input features, y-axis: dim. of hidden nodes)}
        \label{fig:mnist_init_s}
    \end{subfigure}
    \hfill
    \begin{subfigure}[b]{0.48\textwidth}\centering
        \includegraphics[width=\textwidth]{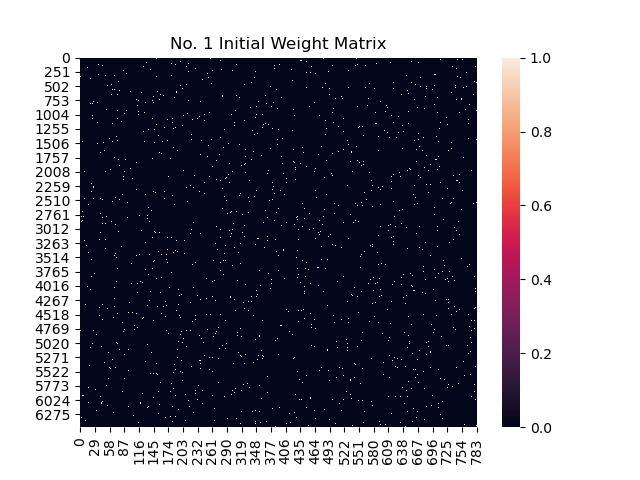}
        \caption{Initial weight matrix of the KCNet with 6,500 hidden nodes before the DOA (x-axis: dim. of input features, y-axis: dim. of hidden nodes)}
        \label{fig:mnist_init_w}
    \end{subfigure}
    \hfill
    \begin{subfigure}[b]{0.48\textwidth}\centering
        \includegraphics[width=\textwidth]{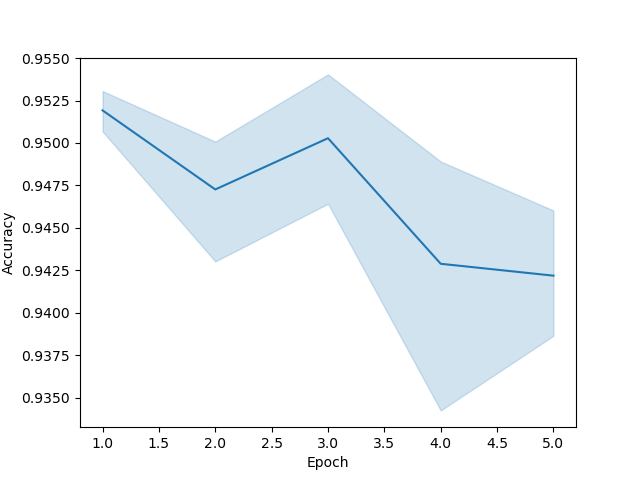}
        \caption{Validation accuracy during the DOA with 5 epochs (x-axis: \# of epochs, y-axis: validation accuracy)}
        \label{fig:mnist_doa_val_acc}
    \end{subfigure}
    \hfill
    \begin{subfigure}[b]{0.48\textwidth}\centering
         \includegraphics[width=\textwidth]{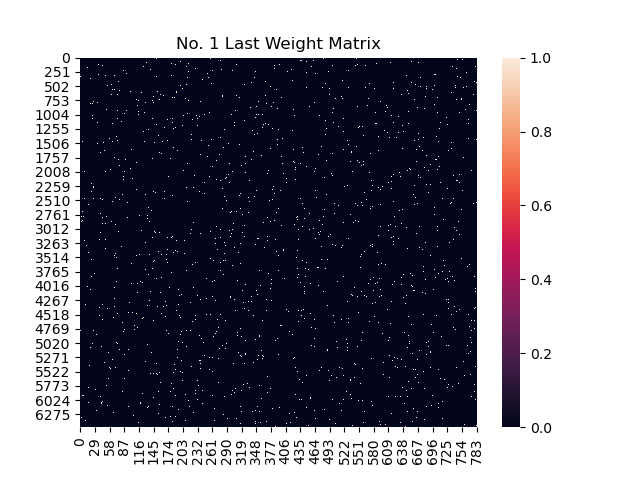}
        \caption{Final weight matrix of the DOA-applied KCNet with 6,500 hidden nodes (x-axis: dim. of input features, y-axis: dim. of hidden nodes)}
        \label{fig:mnist_final_w}
    \end{subfigure}
    \hfill
    \begin{subfigure}[b]{0.48\textwidth}\centering
         \includegraphics[width=\textwidth]{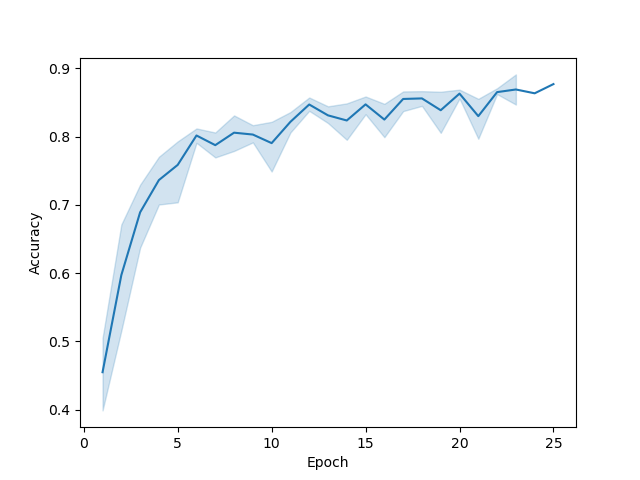}
         \caption{Validation accuracy of 10 submodels during the Ensemble DOA with 100 epochs and \texttt{stop\_metric} = 0.87 (x-axis: \# of epochs, y-axis: validation accuracy)}
         \label{fig:mnist_ensemble_doa_val_acc}
    \end{subfigure}
    \hfill
    \begin{subfigure}[b]{0.48\textwidth}\centering
         \includegraphics[width=\textwidth]{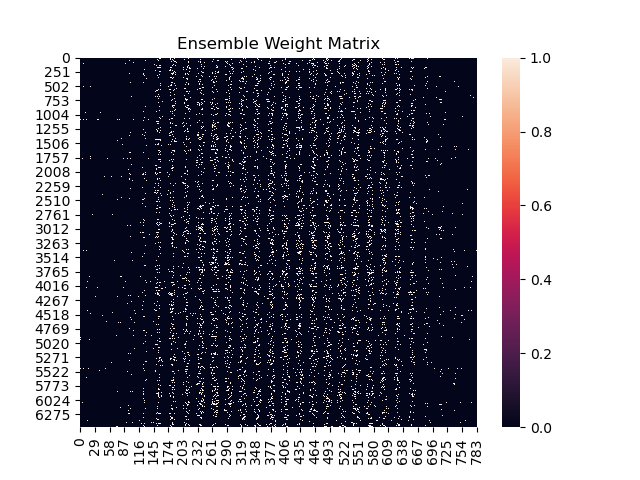}
         \caption{Final weight matrix of the Ensemble-DOA-applied-KCNet with 6,500 hidden nodes (x-axis: dim. of input features, y-axis: dim. of hidden nodes)}
         \label{fig:mnist_ensemble_last_w}
    \end{subfigure}
    \hfill
\caption{For MNIST, results about the DOA-applied KCNet and the Ensemble-DOA-applied KCNet}
\label{ap:mnist_extra}
\end{figure*}

\begin{figure*}[t!]
\centering
    \hfill
    \begin{subfigure}[b]{0.48\textwidth}\centering
         \includegraphics[width=\textwidth]{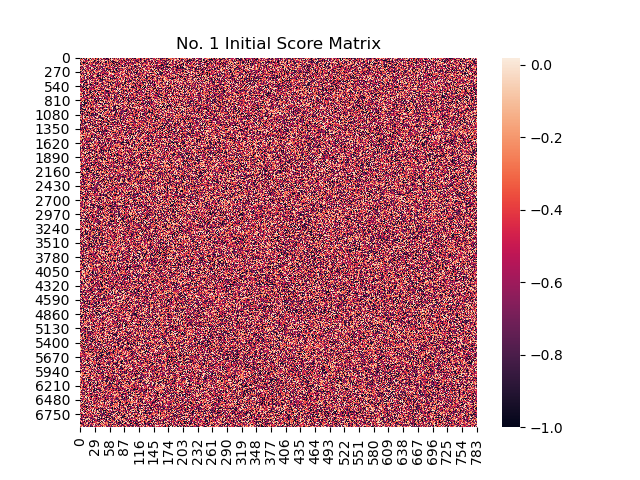}
         \caption{Initial preference score matrix of the KCNet with 7,000 hidden nodes before the DOA (x-axis: dim. of input features, y-axis: dim. of hidden nodes)}
         \label{fig:fashion_mnist_init_s}
    \end{subfigure}
    \hfill
    \begin{subfigure}[b]{0.48\textwidth}\centering
         \includegraphics[width=\textwidth]{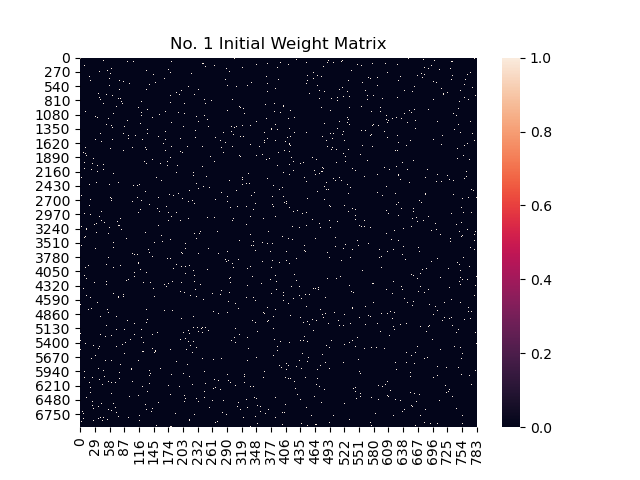}
         \caption{Initial weight matrix of the KCNet with 7,000 hidden nodes before the DOA (x-axis: dim. of input features, y-axis: dim. of hidden nodes)}
         \label{fig:fashion_mnist_init_w}
    \end{subfigure}
    \hfill
    \begin{subfigure}[b]{0.48\textwidth}\centering
         \includegraphics[width=\textwidth]{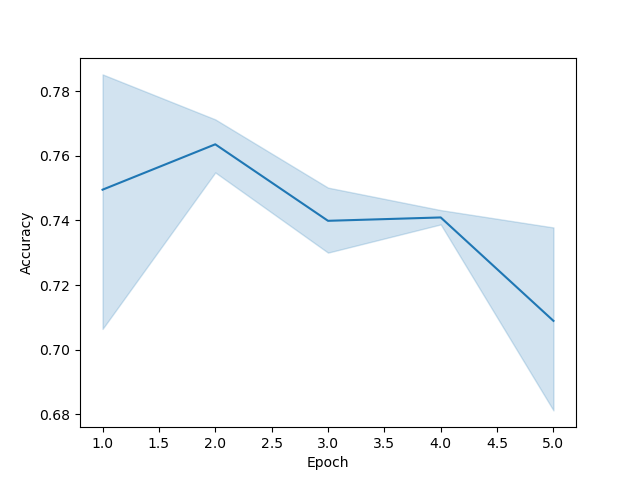}
         \caption{Validation accuracy during DOA with 5 epochs (x-axis: \# of epochs, y-axis: validation accuracy)}
         \label{fig:fashion_mnist_doa_val_acc}
    \end{subfigure}
    \hfill
    \begin{subfigure}[b]{0.48\textwidth}\centering
         \includegraphics[width=\textwidth]{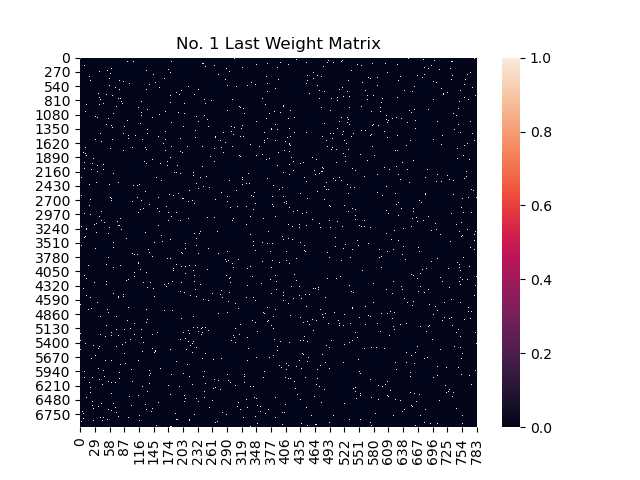}
         \caption{Final weight matrix of the DOA-applied-KCNet with 7,000 hidden nodes (x-axis: dim. of input features, y-axis: dim. of hidden nodes)}
         \label{fig:fashion_mnist_final_w}
    \end{subfigure}
    \hfill
    \begin{subfigure}[b]{0.48\textwidth}\centering
         \includegraphics[width=\textwidth]{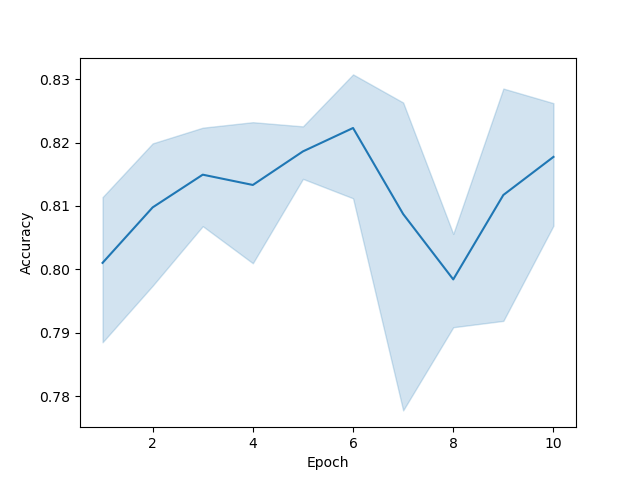}
         \caption{Validation accuracy of 10 submodels during Ensemble DOA with 20 epochs and \texttt{stop\_metric} = 0.84 (x-axis: \# of epochs, y-axis: validation accuracy)}
         \label{fig:fashion_mnist_ensemble_doa_val_acc}
    \end{subfigure}
    \hfill
    \begin{subfigure}[b]{0.48\textwidth}\centering
         \includegraphics[width=\textwidth]{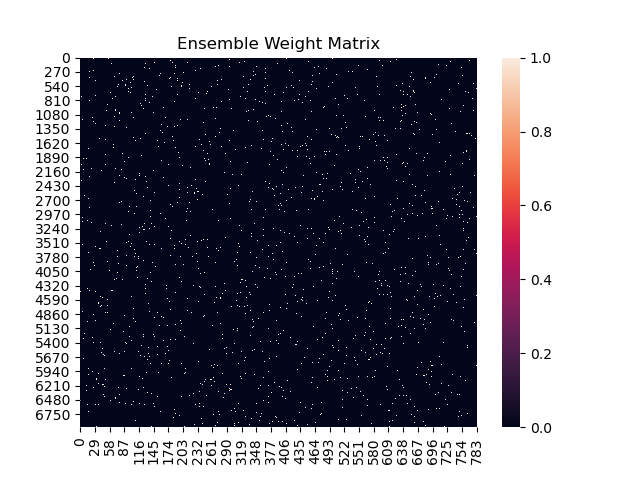}
         \caption{Final weight matrix of the Ensemble-DOA-applied-KCNet with 7,000 hidden nodes (x-axis: dim. of input features, y-axis: dim. of hidden nodes)}
         \label{fig:fashion_mnist_ensemble_last_w}
    \end{subfigure}
    \hfill
    \caption{For Fashion-MNIST, results about the DOA-applied KCNet and the Ensemble-DOA-applied KCNet}
    \label{ap:fashion_mnist_extra}
\end{figure*}

\begin{figure*}[t!]
\centering
\begin{subfigure}[b]{0.48\textwidth}\centering
         \includegraphics[width=\textwidth]{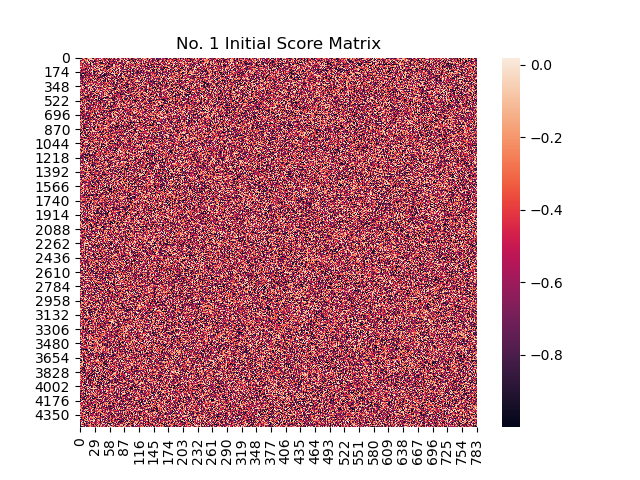}
         \caption{Initial preference score matrix of the KCNet with 4,500 hidden nodes before the DOA (x-axis: dim. of input features, y-axis: dim. of hidden nodes)}
         \label{fig:emnist_init_s}
    \end{subfigure}
    \hfill
    \begin{subfigure}[b]{0.48\textwidth}\centering
         \includegraphics[width=\textwidth]{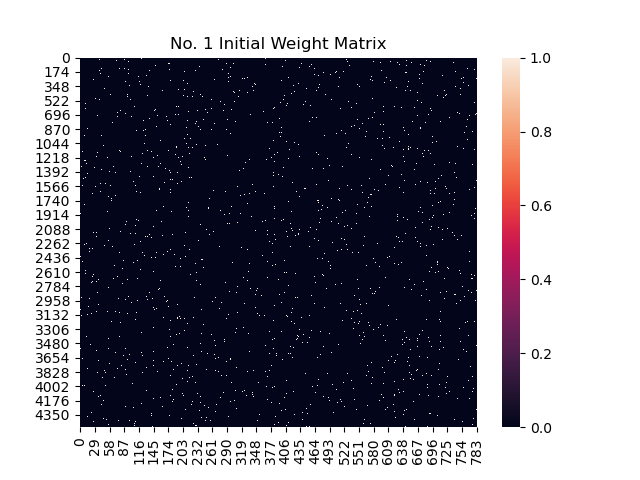}
         \caption{Initial weight matrix of the KCNet with 4,500 hidden nodes before the DOA (x-axis: dim. of input features, y-axis: dim. of hidden nodes)}
     \label{fig:emnist_init_w}
    \end{subfigure}
    \hfill
    \begin{subfigure}[b]{0.48\textwidth}\centering
         \includegraphics[width=\textwidth]{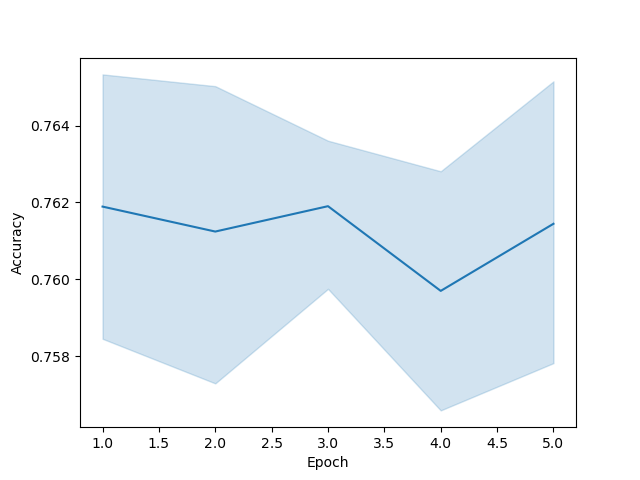}
         \caption{Validation accuracy during DOA with 5 epochs (x-axis: \# of epochs, y-axis: validation accuracy)}
         \label{fig:emnist_doa_val_acc}
    \end{subfigure}
    \hfill
    \begin{subfigure}[b]{0.48\textwidth}\centering
         \includegraphics[width=\textwidth]{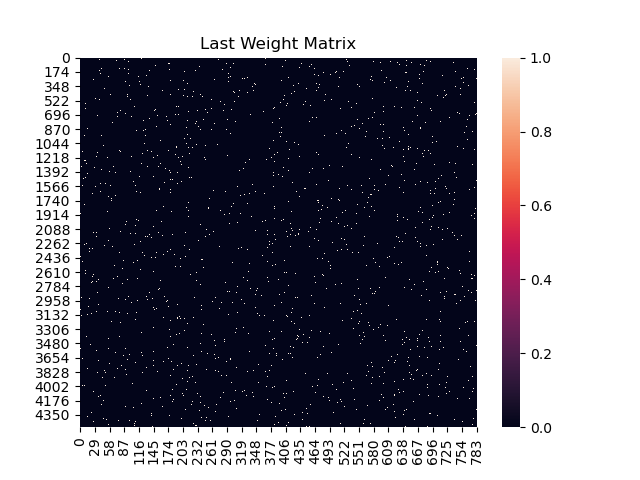}
         \caption{Final weight matrix of the DOA-applied-KCNet with 4,500 hidden nodes (x-axis: dim. of input features, y-axis: dim. of hidden nodes)}
         \label{fig:emnist_final_w}
    \end{subfigure}
    \hfill
    \begin{subfigure}[b]{0.48\textwidth}\centering
         \includegraphics[width=\textwidth]{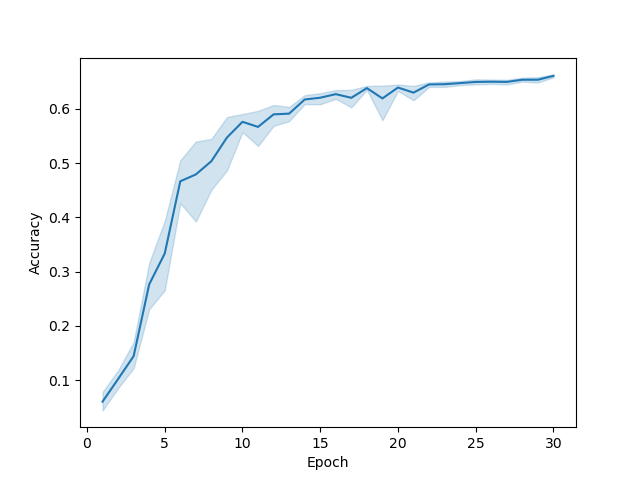}
         \caption{Validation accuracy of 10 submodels during Ensemble DOA with 30 epochs and \texttt{stop\_metric} = 0.85 (x-axis: \# of epochs, y-axis: validation accuracy)}
         \label{fig:emnist_ensemble_doa_val_acc}
    \end{subfigure}
    \hfill
    \begin{subfigure}[b]{0.48\textwidth}\centering
         \includegraphics[width=\textwidth]{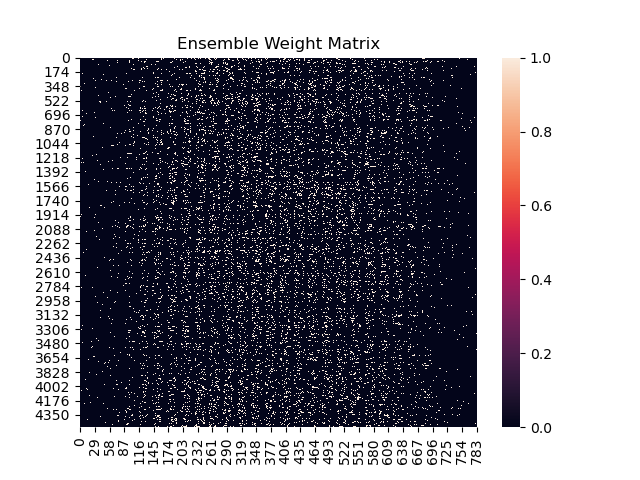}
         \caption{Final weight matrix of the Ensemble-DOA-applied KCNet with 4,500 hidden nodes (x-axis: dim. of input features, y-axis: dim. of hidden nodes)}
         \label{fig:emnist_ensemble_last_w}
    \end{subfigure}
    \hfill
    \caption{For EMNIST-Balanced, results about the DOA-applied KCNet and the Ensemble-DOA-applied KCNet}
    \label{ap:emnist_extra}
\end{figure*}

Figures~\ref{ap:mnist_extra} through~\ref{ap:emnist_extra} depict all the additional results about the DOA-applied KCNet and the Ensemble-DOA-applied KCNet for MNIST, Fashion-MNIST, and EMNIST-Balanced, respectively. We observed that it had a tendency that the DOA couldn't improve the KCNet if its weight matrix was relatively large. As shown in Fig.~\ref{fig:mnist_doa_val_acc} indicating the change of validation accuracy while applying the DOA to the KCNet, the validation accuracy gradually decreased as the DOA progressed. The difference between the weight matrix before~(Fig.~\ref{fig:mnist_init_w}) and after the DOA~(Fig.~\ref{fig:mnist_final_w}) was very marginal, which might be the cause of degrading the performance of the DOA-applied KCNet compared to the original KCNet shown in Table~\ref{tab:result_image}. However, the validation accuracy of 10 submodels during the Ensemble DOA in Fig.~\ref{fig:mnist_ensemble_doa_val_acc} was converged well, and Fig.~\ref{fig:mnist_ensemble_last_w} represents a noticeable improvement of the weight matrix. Therefore, we were able to validate the hyperparameter setting for the Ensemble-DOA-applied KCNet based on the increasing trend of its submodels' validation accuracy, which was applied for other image benchmark datasets. The best hyperparameter settings for the KCNet, the DOA-applied KCNet, and the Ensemble-DOA-applied KCNet for MNIST as follows:
\begin{itemize}
    \item KCNet: \texttt{lambda} = 13
    \item DOA-applied KCNet: \texttt{n\_epoch} = 5, \texttt{learning\_rate} = 1e-4, \texttt{stop\_metric} = 0.96
    \item Ensemble-DOA-applied KCNet : \texttt{n\_epoch} = 100, \texttt{learning\_rate} = 0.01, \texttt{stop\_metric} = 0.87
\end{itemize}

The additional experimental results for Fashion-MNIST~(Fig.~\ref{ap:fashion_mnist_extra}) and EMNIST-Balanced~(Fig.~\ref{ap:emnist_extra}) also tend to be similar to those for MNIST. We specify the best hyperparameter settings for  the KCNet, the DOA-applied KCNet, and the Ensemble-DOA-applied KCNet for Fashion-MNIST as follows:
\begin{itemize}
    \item KCNet: \texttt{lambda} = 5
    \item DOA-applied KCNet: \texttt{n\_epoch} = 5, \texttt{learning\_rate} = 1e-5, \texttt{stop\_metric} = 0.90
    \item Ensemble-DOA-applied KCNet: \texttt{n\_epoch} = 20, \texttt{learning\_rate} = 5e-5, \texttt{stop\_metric} = 0.84
\end{itemize}

The best hyperparameter settings for the KCNet, the DOA-applied KCNet, and the Ensemble-DOA-applied KCNet for EMNIST-Balanced are as follows:
\begin{itemize}
    \item KCNet: \texttt{lambda} = 10
    \item DOA-applied KCNet: \texttt{n\_epoch} = 5, \texttt{learning\_rate} = 1e-4, \texttt{stop\_metric} = 0.90
    \item Ensemble-DOA-applied KCNet: \texttt{n\_epoch} = 30, \texttt{learning\_rate} = 0.01, \texttt{stop\_metric} = 0.85
\end{itemize}

\section{Limitations}
\label{ap:limit}
We showed the potential for our model through the experimental results. However, there are several limitations that we should handle in the future.

In our experiments, we trained and tested our model in supervised batch learning, which has the limitation of scaling the size of the network. This problem has been dealt with in NNRWs, and there are several works about adapting the output weight using online learning~\citep{liang2006fast, van2015online}, which is more practical for large datasets. Online learning also is more congruent with the biological systems that inspired KCNet.

Vulnerability to noisy inputs is the second limit of our model. The main motivation of the KCNet is the transformation of odor stimuli between glomeruli in the antennal lobe and KCs in the mushroom bodies. So our current approach implicitly assumes that, as in the pre-processing of odors in the actual insect's brain, all input features are already pre-processed sufficiently. The image benchmark datasets that we tested satisfied this assumption by applying standard normalization to them, but these conditions are often not applicable to datasets of real-world problems. For example, we tested the KCNet on CIFAR10, an image benchmark dataset with more input features due to the large input and three RGB channels. However, KCNet achieved about 51\% accuracy because the feature dimensions are relatively large, and the purpose of the KCNet is not to distinguish between critical and background features in the images. It instead is a model that generates unique latent representations with combinations of input values. So, directly feeding the real-world images as input into the KCNet may not be a good strategy to take advantage of KCNet's strengths. Just as the central complex~(CX) in the insect's brain is responsible for integrating and elaborating visual information, we anticipate combining an additional feature extractor as an artificial CX, such as pre-trained models, with the KCNet would solve this problem. The KCNet can be a good ad-hoc model for other models, and it will show excellent performance in real-world image tasks if it receives sanitized features processed by a pre-trained model or a simple convolutional network.

\end{document}